\documentclass[lettersize,journal]{IEEEtran}
\usepackage{graphicx}
\usepackage{latexsym}
\usepackage{subfigure}
\usepackage{multirow}
\usepackage{comment}
\usepackage{amsmath}
\usepackage{amsthm}
\usepackage{amssymb}
\usepackage{multirow}
\usepackage{algorithm}
\usepackage{algorithmic}
\usepackage{color}
\usepackage{cite}
\usepackage{enumitem}
\usepackage{booktabs}
\usepackage[pdftex,
bookmarks=true,%
bookmarksnumbered=true,%
colorlinks=true,%
linkcolor=black,%
citecolor=black,%
filecolor=black,%
urlcolor=black%
]{hyperref}

\newcommand{\blue}[1] {\textcolor[rgb]{0.0,0.0,0.0}{{#1}}}  
\newcommand{\ie}{\textit{i}.\textit{e}. }
\newcommand{\ul}[1]{\underline{#1}}

\makeatletter
 
\def\hlinewd#1{%
\noalign{\ifnum0=`}\fi\hrule \@height #1 \futurelet
\reserved@a\@xhline}
\makeatother

\usepackage{microtype}
\usepackage{graphicx}
\usepackage{subfigure}
\usepackage{booktabs} 
\usepackage{hyperref}

\usepackage{amsmath}
\usepackage{amssymb}
\usepackage{mathtools}
\usepackage{amsthm}

\theoremstyle{plain}
\newtheorem{theorem}{Theorem}

\theoremstyle{definition}
\newtheorem{definition}{Definition}

\theoremstyle{remark}

\begin{document}

\title{Efficient Learning of Balanced Signed Graphs via Sparse Linear Programming}

\author{Haruki~Yokota, \IEEEmembership{Member,~IEEE},
        Hiroshi~Higashi, \IEEEmembership{Senior Member,~IEEE},
        Yuichi~Tanaka, \IEEEmembership{Senior Member,~IEEE},
        and~Gene~Cheung,~\IEEEmembership{Fellow,~IEEE}
\thanks{Haruki Yokota and Yuichi Tanaka are with the University of Osaka, Osaka, Japan. (e-mail: hyokota@sip.comm.eng.osaka-u.ac.jp and ytanaka@comm.eng.osaka-u.ac.jp)}
\thanks{Hiroshi Higashi is with Kansai University, Osaka, Japan. (e-mail: hgshrs@kansai-u.ac.jp)}
\thanks{Gene Cheung is with York University, Toronto, Canada. (e-mail:
genec@yorku.ca)}}
\markboth{IEEE Transactions on Signal Processing,~Vol.~X, No.~X, May~2025}%
{Shell \MakeLowercase{\textit{et al.}}: A Sample Article Using IEEEtran.cls for IEEE Journals}


\maketitle

\begin{abstract}
Signed graphs are equipped with both positive and negative edge weights, encoding pairwise correlations as well as anti-correlations in data. 
A balanced signed graph is a signed graph with no cycles containing an odd number of negative edges. 
Laplacian of a balanced signed graph has eigenvectors that map via a simple linear transform to ones in a corresponding positive graph Laplacian, thus enabling reuse of spectral filtering tools designed for positive graphs.
We propose an efficient computation method to learn a balanced signed graph Laplacian directly from data. 
Specifically, extending a previous linear programming (LP) based sparse inverse covariance estimation method called CLIME, we formulate a new LP problem for each Laplacian column $i$, where the linear constraints restrict weight signs of edges stemming from node $i$, so that nodes of same / different polarities are connected by positive / negative edges. 
\blue{We derive a feasible CLIME parameter $\rho_i$ for each sign-constrained column problem.}
We solve the LP problem efficiently by tailoring a sparse LP method based on ADMM. 
\blue{We theoretically prove that the row / column updates produce a non-increasing objective sequence, and show that the iterations are terminated in a finite number of steps.}
Extensive experimental results on synthetic and real-world datasets show that our balanced graph learning method outperforms competing methods and enables reuse of spectral filters, wavelets, and graph neural nets (GNN) constructed for positive graphs.
\end{abstract}


\section{Introduction}
\label{sec:intro}

Modern data with graph-structured kernels can be processed using mathematical tools in \textit{graph signal processing} (GSP) such as graph transforms and wavelets \cite{shuman_emerging_2013,ortega_graph_2018,tanaka_sampling_2020} or deep-learning (DL)-based \textit{graph convolutional nets} (GCN) \cite{kipf2017semi}.  
A basic premise in graph-structured data processing is that a finite graph capturing pairwise relationships is available \textit{a priori}; if such graph does not exist, then it must be learned from observable data---a problem called \textit{graph learning} (GL). 

A variety of GL methods exist in the literature based on statistics, signal smoothness, and diffusion kernels \cite{friedman_sparse_2008a,kalofolias_how_2016,Thanou_learning_2016}. 
However, most methods focus on learning \textit{unsigned} positive graphs, \textit{i.e.}, graphs with positive edge weights that only encode pairwise correlations between nodes. 
As a consequence, most developed graph spectral filters \cite{onuki_graph_2016,pang17,shuman20} and GCNs \cite{kipf2017semi} are also applicable only to positive graphs.
This is understandable, as the notion of \textit{graph spectrum} is well studied for positive graphs---\textit{e.g.}, eigen-pairs $(\lambda_i,\mathbf{v}_i)$ of the combinatorial graph Laplacian matrix $\mathbf{L}$ are commonly interpreted as graph frequencies and Fourier modes, respectively \cite{shuman_emerging_2013,tanaka_sampling_2020,hammond2011wavelets}---and spectral graph filters with well-defined frequency responses are subsequently designed for signal restoration tasks such as denoising, interpolation, and dequantization \cite{gadde2013bilateral,chen2015signal,liu2019graphbased,yazaki2019interpolation,nagahama2022graph}. 

In many practical real-world scenarios, however, there exist datasets with strong pairwise \textit{anti-correlations}.
An illustrative example is voting records in the US Senate, where Democrats / Republicans typically cast opposing votes, and thus edges between them are more appropriately modeled using \textit{negative} weights. 
The resulting structure is a \textit{signed graph}---a graph with both positive and negative edge weights \cite{hou_laplacian_2003,wu_spectral_2011,kunegis_spectral_2010,dittrich_signal_2020}. 
However, the spectra of signed graph variation operators, such as adjacency and Laplacian matrices, are not well understood in general; for example, the nodal domain theorem \cite{davies2000discrete} that shows graph Laplacian eigenvectors of increasing eigenvalues have non-decreasing number of nodal domains (\textit{i.e.}, larger variations across kernel and hence higher frequencies) applies \textit{only} to Laplacian matrices of positive graphs.
Thus, designing spectral filters for general signed graphs remains difficult.

\begin{figure}[tb]
\centering
\includegraphics[width = 0.46\textwidth]{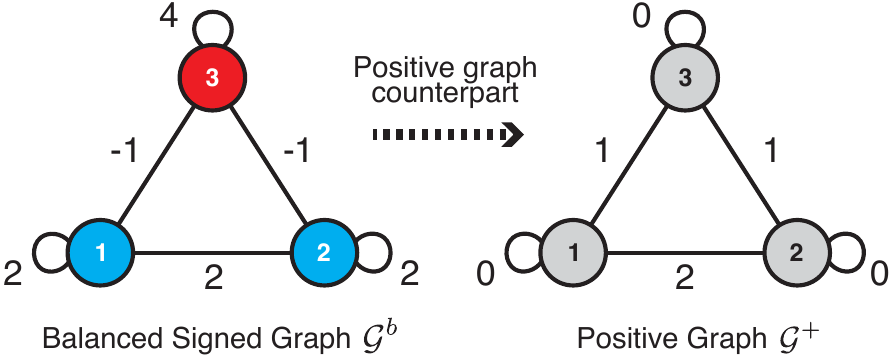}
\vspace{-0.1in}
\caption{Balanced signed graph (Left) and its positive graph counterpart (Right). Light-/dark-colored nodes have positive/negative polarity, respectively. }
\label{fig:balanced_vs_positive}
\end{figure}

One exception is \textit{balanced} signed graphs. 
A signed graph $\mathcal{G}$ is balanced (denoted by $\mathcal{G}^b$) if there exist no cycles with an odd number of negative edges (the formal definition is introduced in Section\;\ref{subsec:balanced_graphs}). 
It is recently discovered that there exists a simple one-to-one mapping from eigenvectors $\mathbf{U}$ of a balanced signed graph Laplacian matrix $\mathcal{L}^b$ to eigenvectors $\mathbf{V} = \mathbf{T} \mathbf{U}$ of a Laplacian $\mathcal{L}^+$ for a corresponding positive graph $\mathcal{G}^+$ (Fig.~\ref{fig:balanced_vs_positive} Right) \cite{Dinesh2025}, where $\mathbf{T}$ is a diagonal matrix with entries $T_{i,i} \in \{1,-1\}$.
Thus, the spectrum of a balanced signed graph $\mathcal{G}^b$ is equivalent to the spectrum of the corresponding positive graph $\mathcal{G}^+$, and well-studied processing tools such as spectral filters for positive graphs can be readily applied to balanced signed graphs.

However, existing GL methods computing balanced signed graphs from data are limited to sub-optimal two-step approaches\footnote{The lone exception is a recent work \cite{matz23}, which employs a complex definition of signed graph Laplacian requiring the absolute value operator \cite{dittrich_signal_2020}. We compare against this work in performance in Section\;\ref{sec:results}.}: first compute a signed graph from data using, for example, \textit{graphical lasso} (GLASSO) \cite{mazumder_graphical_2012}, then balance the computed signed graph via often computation-expensive balancing algorithms
\cite{akiyama_balancing_1981,dinesh_lineartime_2022,yokota_signed_2023}.

In this paper, we propose a computationally-efficient GL method that computes a balanced signed graph $\mathcal{G}^b$ \textit{directly} from observable data.
Specifically, we extend a previous linear programming (LP) based sparse inverse covariance estimation method called CLIME \cite{Cai_aconstrained_2011} to compute a balanced signed graph Laplacian $\mathcal{L}^b$ given a sample covariance matrix $\mathbf{C}$. 
By the \textit{Cartwright-Harary Theorem} (CHT) \cite{harary_notion_1953}, nodes of a balanced signed graph $\mathcal{G}^b$ can be assigned \textit{polarities} $\{1, -1\}$, so that negative / positive edges connect node-pairs of opposing / same polarities, respectively. 
See Fig.\;\ref{fig:balanced_vs_positive}(left) for an illustration of a balanced signed graph with nodes assigned with suitable polarities, denoted by color blue ($1$) and red ($-1$).
Given CHT, the key idea is to impose additional linear sign constraints in the LP formulation on edge weights $w_{i,j}, \forall j$, when node $i$ takes on an assumed polarity during optimization to maintain graph balance.

We solve the resulting \textit{sparse linear programming} (SLP) problem for each column $\mathbf{l}_i$ of $\mathcal{L}^b$ representing node $i$'s connectivity efficiently---linear time $\mathcal{O}(N)$ in node count $N$ when observations $K \ll N$---by tailoring an \textit{alternating direction method of multipliers} (ADMM)-based method \cite{wang17}. 
\blue{For each $\mathbf{l}_i$, we derive a feasible CLIME parameter $\rho_i$ that accounts for the added sign constraints.}
\blue{We theoretically prove that the objectives of our proposed subproblem is non-increasing, and the proposed iterative algorithm terminates in finite steps. }
Experiments show that our method constructs better quality signed graph Laplacians than previous schemes \cite{dinesh_lineartime_2022,yokota_signed_2023,matz23}, and our method enables reuse of previously designed graph filters for positive graphs \cite{pesenson_variational_2009,hammond2011wavelets} and GCNs \cite{kipf2017semi,velickovic2018graph} for signal denoising and interpolation.

The outline of the paper is as follows.
We first overview related work in graph learning in Section\;\ref{sec:related}.
We review basic definitions in GSP, CHT for balanced signed graphs, and sparse inverse covariance matrix learning in Section\;\ref{sec:prelim}.
We formulate our balanced signed graph optimization in Section\;\ref{sec:balanced} and present our algorithm in Section\;\ref{sec:opt}.
Experimental results and conclusion are presented in Section\;\ref{sec:results} and \ref{sec:conclude}, respectively.

\vspace{0.05in}
\noindent
\textbf{Notation:}
Vectors and matrices are written in bold lowercase and uppercase letters, respectively.
The $(i,j)$ element 
of a matrix $\mathbf{A}$ 
is denoted by $A_{i,j}$. 
The $i$-th element in the vector $\mathbf{a}$ is denoted by $a_{i}$.
The square identity matrix of rank $N$ is denoted by $\mathbf{I}_N$, the $M$-by-$N$ zero matrix is denoted by $\mathbf{0}_{M,N}$, and the vector of all ones / zeros of length $N$ is denoted by $\mathbf{1}_N$ / $\mathbf{0}_N$, respectively.
Operator $\|\cdot\|_p$ denotes the $\ell$-$p$ norm. 
$\mathbf{A} \odot \mathbf{B}$ for two matrices of the same dimension denotes the element-by-element multiplication operation.

\section{Related Works}
\label{sec:related}

We first divide our discussion of related works in graph learning into two subsections: i) graph learning from a statistical perspective; and ii) graph learning from a deterministic perspective.
We then overview related works in spectral analysis of (balanced) signed graphs.

\subsection{Statistical Graph Learning}

In this category, an observation matrix $\mathbf{X} = [\mathbf{x}_1, \ldots, \mathbf{x}_K] \in \mathbb{R}^{N \times K}$ composed of $K$ signal observations is available to compute a reasonably reliable sample covariance matrix $\mathbf{C} = \frac{1}{K-1}\mathbf{X} \mathbf{X}^\top$. 
Given $\mathbf{C}$, GLASSO \cite{mazumder_graphical_2012} estimates a sparse inverse covariance (precision) matrix, via the following formulation:
\begin{align}
\min_{\boldsymbol{\Theta} \succeq 0} \text{tr}(\mathbf{C} \boldsymbol{\Theta}) - \log \det (\boldsymbol{\Theta}) + \mu \sum_{j \neq k} |\Theta_{i,j} | 
\label{eq:glasso}
\end{align}
where $\mu > 0$ is a weight parameter for the $\ell_1$-norm signal prior that promotes sparsity in solution $\boldsymbol{\Theta}$. 
The first two terms in \eqref{eq:glasso} can be interpreted as the negative log likelihood.

Extensions of GLASSO abound: \cite{egilmez_graph_2017} constrained solutions $\boldsymbol{\Theta}$ to be sets of targeted categories of graph Laplacian matrices (\textit{e.g.}, combinatorial Laplacians); \cite{Kumar2019} constrained the patterns of the Laplacian eigenvalues so that the obtained graph has desirable properties; \cite{bagheri2024} restricted solutions $\boldsymbol{\Theta}$ to contain a pre-chosen set of core eigenvectors. 
While we also pursue a statistical approach, instead of GLASSO, we start from the linear programming (LP) based CLIME formulation \cite{Cai_aconstrained_2011}, which is more amenable to new constraints that dictate balanced signed graph Laplacian matrices, as we show in Section\;\ref{sec:balanced}.

\subsection{Deterministic Graph Learning}

In scenarios where sufficient signal observations are not available to obtain a reliable sample covariance $\mathbf{C}$, a graph can still be estimated based on additional assumptions.
One popular assumption is that the limited observed signals are all smooth w.r.t. the underlying graph \cite{kalofolias_how_2016,dong16,dong2019}.
For example, \cite{dong16} employed the \textit{graph Laplacian regularizer} (GLR) \cite{pang17}, $\text{tr}(\mathbf{X}^\top \mathbf{L} \mathbf{X})$---a signal smoothness measure for $\mathbf{X}$ given graph specified by $\mathbf{L}$---to regularize an ill-posed problem when seeking Laplacian $\mathbf{L}$. 
However, the assumption on individual signal smoothness may be too strong for some scenarios.

Another popular assumption is the diffusion model \cite{Thanou_learning_2016}: in a time series of signals $\mathcal{X} = \{\mathbf{x}_0, \mathbf{x}_1, \ldots\}$, the signals evolve following differential equation $\frac{\partial \mathbf{x}}{\partial t} = \tau \mathbf{L} \mathbf{x}$, resulting in solution $\mathbf{x}(t) = e^{- \tau \mathbf{L}} \; \mathbf{x}_0$, where $\mathbf{x}_0$ is the initial signal at time $t=0$.
\cite{Thanou_learning_2016} proposed an efficient algorithm to compute $\mathbf{L}$ given $\mathcal{X}$ with an additional sparsity assumption.
Clearly, diffusion models do not apply in scenarios where observed data $\mathcal{X}$ are not evolving signals over time.
Further, even for data in a time series, it is not clear that the pairwise relationships would remain static, resulting in a single fixed graph structure \cite{Bagheri2024icassp}.

Finally, there exist graph learning methods where the assumption is that relevant feature vector $\mathbf{f}_i$ can be computed for each node $i$, so that edge weights of a similarity graph can be computed as function of feature distances \cite{Hu2020,yang_signed_2022}.
Specifically, weight $w_{i,j}$ of edge $(i,j)$ can be computed as
\begin{align}
w_{i,j} &= \exp \left( - d_{i,j} \right) 
\nonumber \\
d_{i,j} &= (\mathbf{f}_i - \mathbf{f}_j)^\top \mathbf{M} (\mathbf{f}_i - \mathbf{f}_j)
\label{eq:featureDist}
\end{align}
where $d_{i,j}$ is the squared feature (Mahalanobis) distance, $\mathbf{f}_i \in \mathbb{R}^K$ is a $K$-dimensional feature vector, and $\mathbf{M} \succeq 0$ is a \textit{positive semi-definite} (PSD) \textit{metric matrix}. 
Feature vectors $\mathbf{f}_i$ can be either hand-crafted or learned directly from data \cite{Do2024}, and the optimization of $\mathbf{M}$ is called \textit{metric learning} \cite{yang_signed_2022}. 
\textit{Bilateral filter} (BF) \cite{tomasi1998bilateral} in image denoising is an example of \eqref{eq:featureDist}, where $\mathbf{f}_i$ contains a pixel's intensity and 2D grid location, and $\mathbf{M}$ is a diagonal matrix $\mathbf{M} = \mathrm{diag}(1/\sigma_d^2, 1/\sigma_r^2)$. 
One possible direction is to consider both signal statistics in sample covariance $\mathbf{C}$ and features $\{\mathbf{f}_i\}$ when estimating an appropriate similarity graph; we leave this exploration for future work.

\subsection{Spectral Analysis of Signed Graphs}

While most spectral analysis works in GSP are done on positive graphs with non-negative edge weights \cite{shuman_emerging_2013,ortega_graph_2018,tanaka_sampling_2020}, there are recent efforts studying graph frequencies on signed graphs. 
Seminally, \cite{kunegis10} defined a signed graph Laplacian $\mathbf{L}^s \triangleq \mathbf{D}^s - \mathbf{W}$ for a signed graph, where the degree $D^s_{i,i}$ of a node $i$ is the sum of the absolute value weights $|w_{i,j}|$ of edges $(i,j)$ stemming from $i$. 
This ensures a PSD Laplacian $\mathbf{L}^s$ and non-negative eigenvalues interpretable as graph frequencies.
Using this definition, \cite{matz23} proposed new methods to learn a signed graph Laplacian from data. 
However, when given a sample covariance matrix $\mathbf{C}$ computed from observed data, how to learn a (sparse) Laplacian $\mathbf{L}^s$ that best reflects statistics in $\mathbf{C}$ is not straightforward, given the nonlinear absolute value operator.
In our experiments, we demonstrate better performance of our method over \cite{matz23} in different graph learning settings.

Spectral properties of balanced signed graphs were studied in \cite{yang_signed_2022}, where it was proven that the left-ends of Gershgorin discs of a balanced signed graph Laplacian $\mathcal{L}^b$ can be perfectly aligned at its minimum eigenvalue $\lambda_{\min}(\mathcal{L}^b)$ via a similarity transform derived from its first eigenvector $\mathbf{v}_1$. 
\cite{Dinesh2025} proved that there exists a one-to-one mapping of eigenvectors of a balanced graph Laplacian matrix $\mathcal{L}^b$ to eigenvectors of a positive graph Laplacian $\mathcal{L}^+$ via a similarity transform derived from node polarities. 
Thus, graph frequencies of a balanced signed graph $\mathcal{G}^b$ are exactly those of the corresponding positive graph $\mathcal{G}^+$, which is well understood in the GSP literature by way of the nodal domain theorem \cite{davies2000discrete}. 
\cite{Dinesh2025} motivates this work; learning a balanced signed graph Laplacian $\mathcal{L}^b$ from sample covariance $\mathbf{C}$ enables us to reuse spectral filters designed for positive graphs $\mathcal{G}^+$ to process signals on balanced signed graphs $\mathcal{G}^b$. 

\vspace{0.05in}
\noindent
\textbf{Improvements from Conference Version}:
This paper extends from our conference version \cite{Yokota2025} in the following ways:
\begin{enumerate}
\item Instead of an off-the-shelf LP solver with complexity $\mathcal{O}(N^{2.055})$ \cite{jiang_2021_faster}, we solve each SLP problem for a column $\mathbf{l}_i$ in balanced graph Laplacian $\mathcal{L}^b$ via a novel adaptation of ADMM \cite{wang17}, resulting in linear-time complexity $\mathcal{O}(N)$ when $K \ll N$;
\item We derive a feasible CLIME parameter $\rho_i$ for each sign-constrained column problem; 
\item \blue{We theoretically prove that the row / column updates produce a non-increasing objective sequence, and show that the iterations terminate in finite steps under a stated numerical stopping rule.}
\item We conducted extensive experiments on synthetic data and four real-world datasets with non-negligible anti-correlations, showing superiority of our algorithm.
\end{enumerate}

\section{Preliminaries}
\label{sec:prelim}

\begin{table}[t]
\caption{Notations used in algorithm development}
\label{tab:notations}
    \centering
    \begin{tabular}{c|l}
    Notations & Descriptions \\ \hline
    $\mathcal{G}(\mathcal{V},\mathcal{E},\mathbf{W})$ & Graph with nodes $\mathcal{V}$, edges $\mathcal{E}$, adjacency matrix $\mathbf{W}$ \\
    $\mathcal{G}^b$ & Balanced signed graph \\
    $\mathcal{L}^b \in \mathbb{R}^{N \times N}$ & Laplacian matrix for balanced signed graph \\
    $\mathcal{G}^+$ & Positive graph \\
    $\mathcal{L}^+ \in \mathbb{R}^{N \times N}$ & Laplacian matrix for positive graph \\
    $\mathbf{X} \in \mathbb{R}^{N \times K}$ & Observation matrix \\
    $\mathbf{C} \in \mathbb{R}^{N \times N}$ & sample covariance matrix \\
    $\mathbf{l}_i \in \mathbb{R}^N$ & $i$-th column of balanced signed graph Laplacian $\mathcal{L}^b$ \\
    $\mathbf{e}_i \in \mathbb{R}^N$ & $i$-th canonical vector \\
    $\rho \in \mathbb{R}_+$ & CLIME parameter specifying sparsity level \\
    $\beta_i \in \{-1,1\}$ & Polarity of node $i$ \\
    $\mathbf{S}^i \in \mathbb{R}^{N \times N}$ & Diagonal matrix s.t. \blue{$S^i_{j,j} = \left\{ \begin{array}{ll}
    \beta_i \beta_j, & \mbox{for}~ j \neq i \\ 
    -1, & \mbox{o.w.}
    \end{array} \right.$} \\
    \blue{$\mathbf{m} \in \mathbb{R}^N$} & \blue{Edge-magnitude variable, $\mathbf{l}_i = -\mathbf{S}^i \mathbf{m}$ with $\mathbf{m} \geq \mathbf{0}_N$} \\
    $\mathbf{r} \in \mathbb{R}^N$ & Dependent variable $\mathbf{r} \triangleq \mathbf{C} \mathbf{l}_i - \mathbf{e}_i$ \\
    $\mathbf{q} \in \mathbb{R}^{\blue{2N}}$ & Non-negative slack variable \\
    \blue{$\mathbf{y} \in \mathbb{R}^{4N}$} & \blue{Stacked variable $\mathbf{y} = [\mathbf{m}; \mathbf{r}; \mathbf{q}]$ in \eqref{eq:LP}} \\
    $\mathbf{c} \in \mathbb{R}^{\blue{4N}}$ & Cost vector \blue{$\mathbf{c} \triangleq [\mathbf{1}_N; \mathbf{0}_{3N}]$} in \eqref{eq:LP} \\
    \blue{$\mathbf{E} \in \{0,1\}^{3N \times 4N}$} & \blue{Selection matrix, $\blue{\mathbf{z}} = [\mathbf{m}; \mathbf{q}] = \mathbf{E}\mathbf{y}$} \\
    $\tilde{\mathbf{z}} \in \mathbb{R}^{\blue{3N}}$ & Auxiliary variable \\
    $\boldsymbol{\mu} \in \mathbb{R}^{\blue{6N}}$ & Lagrange multiplier \\
    $\gamma \in \mathbb{R}_+$ & ADMM parameter\\\hline
    \end{tabular}\vspace{-10pt}
\end{table}

We first formally define signed graphs and balanced signed graphs.
We then review a sparse inverse covariance matrix estimation algorithm called CLIME \cite{Cai_aconstrained_2011}.
See Table\;\ref{tab:notations} for a list of notations.

\subsection{Signed Graphs}
\label{sec:signed_graphs}

Denote by $\mathcal{G} (\mathcal{V},\mathcal{E},\mathbf{W})$ an undirected signed graph with node set $\mathcal{V} = \{1, \dots, N\}$, edge set $\mathcal{E}$, and symmetric weighted adjacency matrix $\mathbf{W} \in \mathbb{R}^{N \times N}$, where weight of edge $(i,j) \in \mathcal{E}$ connecting nodes $i, j \in \mathcal{V}$ is $w_{i,j} = W_{i,j} \in \mathbb{R}$. 
We assume that $W_{i,j}, \forall i\neq j$, can be positive / negative to encode pairwise correlation / anti-correlation, while (possible) self-loop weight $w_{i,i}$ is non-negative, \textit{i.e.}, $W_{i,i} \geq 0, \forall i$. 
Denote by  $\mathbf{D} \in \mathbb{R}^{N \times N}$ a diagonal \textit{degree matrix}, where $D_{i,i} = \sum_{j} W_{i,j}$. 
We define the symmetric \textit{generalized graph Laplacian matrix}\footnote{We use the generalized graph Laplacian for signed graphs instead of the signed graph Laplacian definition in \cite{dittrich_signal_2020} that requires the absolute value operator. This leads to a simple mapping of eigenvectors of our signed graph Laplacian to ones of a corresponding positive graph Laplacian, and a straightforward LP problem formulation extending from \cite{Cai_aconstrained_2011}.} for $\mathcal{G}$ as $\mathcal{L} \triangleq \mathbf{D} - \mathbf{W} + \operatorname{diag}(\mathbf{W})$ \cite{ortega_graph_2018}.

\subsection{Balanced Signed Graphs}
\label{subsec:balanced_graphs}

A signed graph is balanced if there are no cycles with an odd number of negative edges---the product of edge weights in any cycle is positive \cite{harary_notion_1953}. 
We rephrase an equivalent definition of balanced signed graphs, known as \textit{Cartwright-Harary Theorem} (CHT), as follows.
\begin{theorem}
\textit{A signed graph $\mathcal{G}(\mathcal{V}, \mathcal{E}, \mathbf{W})$ is balanced if and only if each node $i \in \mathcal{V}$ can be assigned polarity $\beta_i \in \{1, -1\}$, such that a positive / negative edge always connects nodes of the same / opposite polarities, \ie $\beta_i \beta_j = \text{sign}(W_{i,j})$.}
\end{theorem}
In other words, a signed graph is balanced if all graph nodes can be partitioned into two disjoint sets $\mathcal{V}_1$ and $\mathcal{V}_{-1}$, 
such that all edges \textit{within} a set $\mathcal{V}_k$, $k\in \{1, -1\}$, are positive, and all edges \textit{between} the two sets $\mathcal{V}_1$ and $\mathcal{V}_{-1}$ are negative.
In this case, the edges are said to be \textit{consistent}.
In \cite{Dinesh2025}, the notion of consistent edges in a balanced graph is defined as follows:
\begin{definition}
\label{def:consistent}
\textit{A consistent edge $(i,j) \in \mathcal{E}$ with weight $W_{i,j}$ is a positive / negative edge connecting two nodes of the same / opposite polarities, i.e., $\beta_i \beta_j = \textit{sign}(W_{i,j})$.}
\end{definition}

\subsection{Mapping to Positive Graph Counterpart}

We define the Laplacian $\mathcal{L}^+$ for the positive graph counterpart $\mathcal{G}^+$ of a balanced signed graph $\mathcal{G}^b$ as $\mathcal{L}^+ \triangleq \mathbf{D}-|\mathbf{W}|+\operatorname{diag}(\mathbf{W})$, where $|\mathbf{W}|$ denotes a matrix with element-wise absolute value of $\mathbf{W}$.
A positive graph $\mathcal{G}^+$ has its own adjacency matrix\footnote{The use of a self-loop of weight equaling to twice the sum of connected negative edge weights is also done in \cite{su17}.} $\mathbf{W}^+$, where $W_{i,j}^+ = |W_{i,j}|, \forall i \neq j$, and $W_{i,i}^+ = W_{i,i} - 2 \sum_{j} [-W_{i,j}]_+$, where $[c]_+$ is a positivity function that returns $c$ if $c \geq 0$ and $0$ otherwise.
The diagonal degree matrix $\mathbf{D}^+$ for $\mathcal{G}^+$ is defined conventionally, where $D_{i,i}^+ = \sum_j W_{i,j}^+$.
Note that the adjacency matrix $\mathbf{W}^+$ is so defined such that 
\begin{align}
\mathcal{L}^+ = \mathbf{D} - |\mathbf{W}| + \mathrm{diag}(\mathbf{W}) = \mathbf{D}^+ - \mathbf{W}^+ + \mathrm{diag}(\mathbf{W}^+) .
\end{align}
One can show via the \textit{Gershgorin circle theorem} (GCT) \cite{varga04} that a sufficient condition for any $\mathcal{L}^+$ to be \textit{positive semi-definite} (PSD) is $\mathrm{diag}(\mathbf{W}^+) \geq \mathbf{0}_N$.
For illustration, in Fig.\;\ref{fig:balanced_vs_positive}, we show an example of a balanced signed graph $\mathcal{G}^b$ and its positive graph counterpart $\mathcal{G}^+$. 

A balanced signed graph Laplacian $\mathcal{L}^b = \mathbf{U} \boldsymbol{\Lambda} \mathbf{U}^\top$ of $\mathcal{G}^b$ enjoys a one-to-one mapping of its eigenvectors to those of its positive graph counterpart $\mathcal{L}^+$ \cite{Dinesh2025}.
Specifically, suppose we partition the nodes of a balanced signed graph $\mathcal{G}^b=(\mathcal{V},\mathcal{E})$, into two disjoint sets $\mathcal{V}_1$ and $\mathcal{V}_{-1}$ and we reorder rows / columns of balanced signed graph Laplacian $\mathcal{L}^b$, so that nodes in $\mathcal{V}_{1}$ of size $M$ are indexed before nodes in $\mathcal{V}_{-1}$ of size $N-M$. 
Then, using the following invertible diagonal matrix 
\begin{align}
\label{eq:matrixT}
\mathbf{T} &= \left[ \begin{array}{cc}
\mathbf{I}_M & \mathbf{0}_{M,N-M} \\
\mathbf{0}_{N-M,M} & -\mathbf{I}_{N-M}
\end{array} \right], 
\end{align}
we can show that $\mathcal{L}^b$ and $\mathcal{L}^+$ are \textit{similarity transform} of each other: 
\begin{align}
\mathbf{T} \mathcal{L}^b \mathbf{T}^{-1} &\stackrel{(a)}{=} \mathbf{T} \mathbf{U} \boldsymbol{\Lambda} \mathbf{U}^\top \mathbf{T}^\top \stackrel{(b)}{=} \mathbf{V} \boldsymbol{\Lambda} \mathbf{V}^\top \nonumber
\\
&\stackrel{(c)}{=} \mathbf{T} \left( \mathbf{D} - \mathbf{W} + \mathrm{diag}(\mathbf{W}) \right) \mathbf{T} \\
&= \mathbf{D} - |\mathbf{W}| + \mathrm{diag}(\mathbf{W}) = \mathcal{L}^+\nonumber
\end{align}
where in $(a)$ we write $\mathbf{T}^\top = \mathbf{T}^{-1} = \mathbf{T}$, in $(b)$ $\mathbf{V} \triangleq \mathbf{T} \mathbf{U}$, and in $(c)$ we apply the definition of generalized graph Laplacian.
Thus, $\mathcal{L}^+$ and $\mathcal{L}^b$ have the same set of eigenvalues, and their eigen-matrices are related via matrix $\mathbf{T}$. 
For the example in Fig.\;\ref{fig:balanced_vs_positive}, since the signed graph on the left is balanced, $\mathcal{L}^+$ is a similar transform of $\mathcal{L}^b$ via diagonal matrix $\mathbf{T} = \mathrm{diag}([1,1,-1])$.

\subsection{Spectral Graph Filters} 

Spectral graph filters are frequency modulating filters, where graph frequencies are conventionally defined as the \textit{non-negative} eigenvalues of a PSD graph Laplacian \cite{ortega_graph_2018}. 
Thus, to reuse spectral filters designed for positive graphs in the literature \cite{shuman20} for a balanced signed graph $\mathcal{G}^b$, it is sufficient to require $\mathrm{diag}(\mathbf{W}^+) \geq \mathbf{0}$ for the adjacency matrix $\mathbf{W}^+$ of the positive graph counterpart $\mathcal{G}^+$ \cite{cheung18}.
This condition translates to self-loops of weights $W_{i,i} \geq 2 \sum_j[-W_{i,j}]_+, \forall i$, for the balanced signed graph $\mathcal{G}^b$.

\subsection{Sparse Inverse Covariance Learning}
\label{sec:clime}

Given a data matrix $\mathbf{X}=[\mathbf{x}_1,\dots,\mathbf{x}_K] \in \mathbb{R}^{N \times K}$ where $\mathbf{x}_k \in \mathbb{R}^{N}$ is the $k$th signal observation, a sparse inverse covariance matrix $\mathcal{L} \in \mathbb{R}^{N\times N}$ (interpretable as a signed graph Laplacian) can be estimated from data via a constrained $\ell_1$-minimization formulation called CLIME \cite{Cai_aconstrained_2011}. 
In a nutshell, CLIME seeks $\mathcal{L}$ given a sample covariance matrix\footnote{Sample covariance matrix $\mathbf{C}$ is \textit{positive definite} (PD) if $\mathbf{X}$ is full row-rank, which is possible only if $K \geq N$. We consider both cases, $K \geq N$ and $K \ll N$, in the sequel.} \blue{$\mathbf{C} = \frac{1}{K-1}\mathbf{XX}^\top$} (assuming zero mean) via a \textit{linear programming} (LP) formulation:
\begin{equation}
\label{eq:clime}
\min_\mathcal{L}\quad \|\mathcal{L}\|_1, \quad \text{s.t.}\quad \|\mathbf{C} \mathcal{L}-\mathbf{I}_N\|_\infty\leq \rho
\end{equation}
where $\rho \in\mathbb{R}_+$ is a positive scalar parameter. 
Specifically, \eqref{eq:clime} computes a sparse $\mathcal{L}$ (promoted by the $\ell_1$-norm objective) that approximates the right inverse of $\mathbf{C}$. 
The $i$th column $\mathbf{l}_i$ of $\mathcal{L}$ can be computed independently:
\begin{equation}
\label{eq:clime_i}
\min_{\mathbf{l}_i}\quad \|\mathbf{l}_i\|_1,\quad\text{s.t.}\quad\|\mathbf{C} \, \mathbf{l}_i-\mathbf{e}_i\|_{\infty}\leq\rho 
\end{equation}
where $\mathbf{e}_i$ is the canonical vector with one at the $i$-th entry and zero elsewhere. 
\eqref{eq:clime_i} can be solved using an off-the-shelf LP solver such as \cite{jiang_2021_faster} with complexity $\mathcal{O}(N^{2.055})$. 
The resulting matrix $\mathcal{L} = [\mathbf{l}_1,\mathbf{l}_2,\dots,\mathbf{l}_N]$ is not symmetric in general, and 
a symmetric approximation $\mathcal{L}^* \triangleq (\mathcal{L} + \mathcal{L}^\top) / 2$ \blue{is computed} in a post-processing step\footnote{\blue{Note that post-processing steps such as matrix symmetrization and PSD-enforcing diagonal loading alter the previously optimized columns $\mathbf{l}_i$; hence, the resulting matrix is not guaranteed to satisfy \(\|\mathbf C\mathbf l_i-\mathbf e_i\|_\infty\leq\rho\) for every \(i\).
While one can directly incorporate matrix symmetry and PSDness as linear constraints into \eqref{eq:clime_i}, for computational complexity reasons we retain the simpler formulation \eqref{eq:clime_i} as a starting point.}}. 
In \cite{Dinesh_complex_2023}, CLIME is extended to learn a complex-valued Hermitian graph Laplacian matrix.

\blue{We adopt a CLIME-style formulation, rather than GLASSO \cite{friedman_sparse_2008a}, because the linear edge sign constraints can be added to each column-wise LP while preserving convexity and separability, thereby enabling the sparse ADMM implementation in Section\;\ref{sec:opt}. 
Although the edge sign constraints can also be imposed on GLASSO without losing convexity, its log-determinant term couples the full precision matrix and prevents independent column-wise optimization. }

\section{Baseline Balanced Signed Graph Optimization}
\label{sec:balanced}

\begin{figure*}[tb]
\centering
\includegraphics[width = 0.9\textwidth]{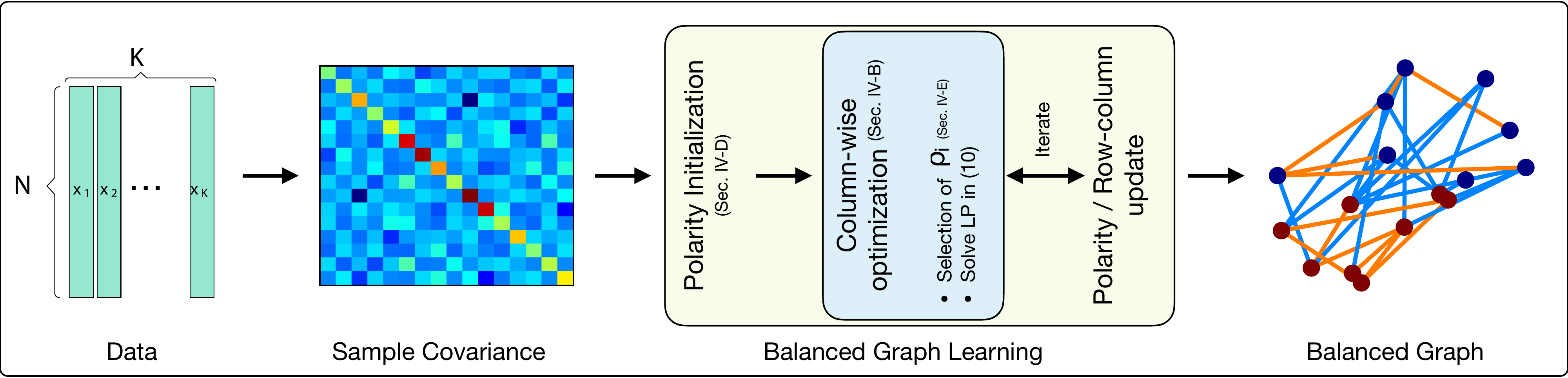}
\caption{Overview of proposed balanced graph learning approach.}\vspace{-15pt}
\label{fig:overview}
\end{figure*}

We first overview our optimization approach to learn a generalized Laplacian $\mathcal{L}^b$ corresponding to a balanced signed graph $\mathcal{G}^b$, given a sample covariance matrix $\mathbf{C}$.
We then formulate an LP-based optimization for each column $\mathbf{l}_i$ of $\mathcal{L}^b$, extended from CLIME \cite{Cai_aconstrained_2011}.
An overview of the proposed method is shown in Fig.\;\ref{fig:overview}.

\subsection{Optimization Approach}
\label{subsec:approach}

Our goal is to estimate a balanced signed graph Laplacian $\mathcal{L}^b\in\mathbb{R}^{N\times N}$ directly from \blue{$\mathbf{C} = \frac{1}{K-1}\mathbf{X} \mathbf{X}^\top$}. 
In essence, our algorithm optimizes polarity $\beta_i \in \{-1,1\}$ and edge weights $w_{i,j}$'s stemming from one node $i \in \{1, \ldots, N\}$ at a time, and terminates when the objective no longer improves. 
As discussed in Section\;\ref{subsec:balanced_graphs}, CHT states that in a balanced signed graph, same- / opposite-polarity node pairs can only be connected by edges of positive / negative weights, respectively. 
Given the edge sign restrictions of a balanced graph dictated by CHT, for each node $i$, we execute a variant of optimization \eqref{eq:clime_i} \textit{twice}---one for each possible polarity $\beta_i$---where the signs of the entries in variable $\mathbf{l}_i$ (weight of edges stemming from node $i$) are restricted for consistency.  
We then adopt node $i$'s polarity $\beta_i$ corresponding to the smaller of the two objective values as the optimal polarity. 

Specifically, our optimization algorithm is as follows:

\vspace{0.05in}
\begin{enumerate}
\item Initialize polarities $\beta_j$ for all graph nodes $j$.
\item For each column $\mathbf{l}_i$ of $\mathcal{L}^b$ corresponding to node $i$,
\begin{enumerate}
    \item Assume node $i$'s polarity $\beta_i \in \{1, -1\}$.
    \item Optimize $\mathbf{l}_i$ while restricting $\text{sign}(l_{i,j})$ so that $l_{i,j}$'s are consistent with polarities of connected nodes $j$.
    \item Smaller of the two objective values corresponding to assumptions $\beta_i = 1$ and $\beta_i = -1$ determines node $i$'s polarity.
    \item Update $i$th column/row of $\mathcal{L}^b$ using computed $\mathbf{l}_i$.
\end{enumerate}
\item Continue updating columns of $\mathcal{L}^b$ \blue{till there is no change in objective value for all columns $\mathbf{l}_i$'s}.
\end{enumerate}
\vspace{0.05in}
Note that by simultaneously updating the $i$th column / row of $\mathcal{L}^b$ in step 2(d), $\mathcal{L}^b$ remains symmetric. 

An alternative approach is to alternately optimize edge weight magnitude in  $\mathcal{L}^b$ and the node-polarity matrix $\mathbf{B}=\boldsymbol{\beta\beta}^\mathsf{T}$ for $\boldsymbol{\beta} = [\beta_1,\dots\beta_N]$, where the optimization of $\mathbf{B}$ is formulated as a \textit{semi-definite programming} (SDP) problem \cite{Luo2010}. However, computation of a general SDP is expensive (at least $\mathcal{O}(N^3)$). Further, the initial polarities extracted from the covariance matrix (described in Section \ref{subsec:initialization}) have proven sufficiently accurate, and thus few further updates are necessary.




\subsection{Linear Constraints for Consistent Edges}

To reformulate CLIME for a balanced signed graph Laplacian $\mathcal{L}^b$, we first write linear constraints that ensure consistent edges. 
From Definition \ref{def:consistent}, we rewrite the edge consistency restriction $\beta_i \beta_j = \text{sign}(W_{i,j})$ for an edge $(i,j)\in \mathcal{E}$ of the target balanced graph $\mathcal{G}^b$, given node polarities $\beta_i$ and $\beta_j$, as
\begin{equation}
\label{eq:consistent}
    \beta_i\beta_j\operatorname{sign}(W_{i,j})=
\begin{cases}1 & \text{if }(i,j)\text{ is consistent}\\-1 & \text{if }(i,j)\text{ is inconsistent}\end{cases} .
\end{equation}
For each Laplacian entry $\mathcal{L}_{i,j}^b$, linear constraint for edge consistency, given node polarities $\beta_i$ and $\beta_j$, is thus
\begin{equation}
\label{eq:b_const}
    \beta_i\beta_j \mathcal{L}_{i,j}^b \leq 0, ~~~\forall j \,|\, j \neq i .
\end{equation}

\subsection{Optimization of Laplacian Column}
\label{sec:prop}

We can now reformulate the LP problem \eqref{eq:clime_i} by incorporating linear constraints \eqref{eq:b_const}.
Denote by $\mathbf{S}^i \in\mathbb{R}^{N \times N}$ a diagonal matrix for optimization of $\mathbf{l}_i$, where \blue{the diagonal entries are $S_{j,j}^i = \beta_i \beta_j$ for $j\neq i$, and $S_{i,i}^i=-1$}.
Optimization \eqref{eq:clime_i} becomes
\begin{equation}
\label{eq:prop2}
\min_{\mathbf{l}_i} \|\mathbf{l}_i\|_1 \quad\text{s.t.} \quad \begin{cases} \|\mathbf{C} \mathbf{l}_i-\mathbf{e}_i \|_\infty\leq\rho_i\\\mathbf{S}^i \mathbf{l}_{i}\leq \mathbf{0}_N \end{cases} .
\end{equation}
The vector inequality here represents entry-wise inequality. 
By incorporating one more linear constraint, \eqref{eq:prop2} remains an LP. Note that the original formulation \eqref{eq:clime_i} assumes a single parameter $\rho$ for all columns. However, there is no guarantee that each column enjoys the same sparsity. So in our formulation, we adaptively set $\rho_i$ for each column-wise LP.

\vspace{0.1in}
\subsubsection{Feasibility of \eqref{eq:prop2}}

Unlike \eqref{eq:clime_i} that always has at least one feasible solution for any $\rho > 0$ (\textit{i.e.}, $i$th column of $\mathbf{C}^{-1}$, assuming $K \geq N$), the additional inequality constraint in \eqref{eq:prop2} means that the LP may be \textit{infeasible}. 
Thus, the selection of $\rho_i$ is particularly crucial for \eqref{eq:prop2}; we discuss this issue in Section\;\ref{subsec:rho}.
Assuming that the LP \eqref{eq:prop2} is feasible for every $i$, we prove \blue{finite-step termination} of our iterative algorithm for $\mathcal{L}^b$ in Section\;\ref{subsec:convergence}. 
Instead of using an off-the-shelf solver like \cite{jiang_2021_faster} designed for general LP, we tailor an ADMM-based method \cite{wang17} for our particular LP \eqref{eq:prop2} in Section\;\ref{sec:opt}, which has a speed advantage for the $K \ll N$ case.

\vspace{0.1in}
\subsubsection{Mismatch of Statistical Models}
\label{subsec:mismatch}
In general, the statistics (inverse covariance $\mathbf{C}^{-1}$) of a given dataset may not be a balanced signed graph Laplacian.
Thus, by proactively enforcing graph balance in \eqref{eq:prop2}, the computed Laplacian $\mathcal{L}^b$ may suffer from mismatch to the true statistics of the data.
However, there are ample empirical evidence that typical datasets in practice are \blue{\textit{nearly balanced}}, including social, biological, economic, and collaborative networks \cite{antal2005dynamic,hu2013bipartite,rawlings2017structural,gallo2024testing}. 
\blue{
The benefit of a strong assumption that the underlying graph is \textit{perfectly balanced} is that spurious (anti-)correlations due to insufficient and/or noisy data can be eliminated. 
This is analogous to the \textit{piecewise smooth} (PWS) assumption common in image restoration \cite{cai2016image,ji2016image}. 
Although natural images are not perfectly PWS, the assumption enables mathematically convenient signal priors such as \textit{total variation} (TV) \cite{liu2019image}, resulting in superior restoration performance such as denoising.}
Similarly, in our experiments in Section\;\ref{sec:results} we show that for many real-world datasets, the benefits of learning a balanced signed graph  outweigh the cost of potential statistical mismatches.

\subsection{Initial Nodal Polarity Assignments}
\label{subsec:initialization}

As discussed in Section\;\ref{subsec:approach}, initialization of node polarities $\{\beta_i\}$ is an important first step.
\blue{We initialize polarities using the signs of the dominant eigenvector of the sample covariance matrix $\mathbf{C} = \tfrac{1}{K-1}\mathbf{X}\mathbf{X}^\top$. 
Denote by $\mathbf{u}_1$ the eigenvector corresponding to the largest eigenvalue of $\mathbf{C}$.
We assign
\begin{align}
    \beta_i = \operatorname{sign}\!\left((\mathbf{u}_1)_i\right),
    \qquad i = 1, \ldots, N ,
    \label{eq:eig_init}
\end{align}
with a fixed convention for zero entries\footnote{Since eigenvectors are identifiable only up to a global sign, $\boldsymbol{\beta}$ and $-\boldsymbol{\beta}$ define the same balanced partition.}.}

\blue{This initialization is exact in the ideal case.
Suppose that the sign pattern of $\mathbf{C}$ is consistent with a balanced graph with polarities $\boldsymbol{\beta}^\star$, \ie
$\operatorname{sign}(C_{i,j}) = \beta_i^\star \beta_j^\star$ for all nonzero entries.
Then, the similarity transform $\mathbf{M} = \operatorname{diag}(\boldsymbol{\beta}^\star) \, \mathbf{C}  \operatorname{diag}(\boldsymbol{\beta}^\star)$ is entry-wise nonnegative with the same eigenvalues as $\mathbf{C}$.
If $\mathbf{M}$ is additionally irreducible, then the Perron--Frobenius theorem guarantees a strictly positive leading eigenvector $\mathbf{v}_1$ of
$\mathbf{M}$, and hence $\mathbf{u}_1 = \operatorname{diag}(\boldsymbol{\beta}^\star)\mathbf{v}_1$ satisfies $\operatorname{sign}(\mathbf{u}_1) = \pm\boldsymbol{\beta}^\star$. 
Thus, \eqref{eq:eig_init} recovers the true polarity pattern whenever the covariance sign pattern is exactly balanced, and degrades gracefully as the pattern deviates from balance.}

\blue{We approximate $\mathbf{u}_1$ with $m$ Lanczos iterations\footnote{Note that any iterative method to estimate the dominant eigenvector would suffice, such as LOBPCG \cite{knyazev2001lobpcg}.}. 
Starting from a unit vector $\mathbf{q}_1$ with $\mathbf{q}_0=\mathbf{0}$ and $\beta_0=0$, iteration $j$ computes 
\begin{align}
\begin{split}
    \mathbf{z}_j &=
    \frac{1}{K-1}\mathbf{X}(\mathbf{X}^{\top}\mathbf{q}_j)
    -\beta_{j-1}\mathbf{q}_{j-1}, \nonumber\\
    \alpha_j&=\mathbf{q}_j^\top\mathbf{z}_j,\quad
    \mathbf{z}_j\leftarrow\mathbf{z}_j-\alpha_j\mathbf{q}_j,\\
    \beta_j&=\|\mathbf{z}_j\|_2,\quad
    \mathbf{q}_{j+1}=\mathbf{z}_j/\beta_j.
\end{split}
\end{align}
Let $\mathbf{Q}_m=[\mathbf{q}_1,\ldots,\mathbf{q}_m]$ and let $\mathbf{T}_m$ be the symmetric tridiagonal matrix with diagonals $\{\alpha_j\}_{j=1}^m$ and off-diagonals $\{\beta_j\}_{j=1}^{m-1}$.
If $\mathbf{y}_1$ is the dominant eigenvector of $\mathbf{T}_m$, then the leading vector $\widehat{\mathbf{u}}_1=\mathbf{Q}_m\mathbf{y}_1$
approximates $\mathbf{u}_1$.}

\subsection{Choosing CLIME Parameter $\rho_i$}
\label{subsec:rho}

\blue{The parameter $\rho_i$ in \eqref{eq:prop2} controls the covariance-fit tolerance and resulting sparsity.
Information criteria such as AIC, BIC, and HQIC \cite{akaike1998aic,schwarz1978bic,hannan1979hqic} can rank candidate values of $\rho$ based on model fit and complexity.
However, applying them here requires a symmetric positive-definite estimate; moreover, they could be unreliable when $K\ll N$ and do not ensure column-wise feasibility.
We therefore compute the minimum feasible value $\rho^l$ and set
$\rho_i=\rho^l+\delta_0$, where $\delta_0>0$ is a small numerical
margin. These steps are described next.}

\vspace{0.05in}
\subsubsection{\textbf{Choosing $\rho_i$ for Minimum Feasibility}}

For the first time LP \eqref{eq:prop2} is executed for a particular node $i$, we compute $\rho^l$ to ensure \eqref{eq:prop2} has at least a feasible solution $\mathbf{l}_i$. 
To accomplish this, we compute the following LP \blue{over $\mathbf l_i$ and $\rho$:}
\begin{align}
    \rho^l
    =
    \min_{\mathbf l_i,\rho}\quad
    &\rho
    \nonumber\\
    \mathrm{s.t.}\quad
    &\mathbf e_i-\rho\mathbf 1_N
    \leq \mathbf C\mathbf l_i
    \leq \mathbf e_i+\rho\mathbf 1_N,\nonumber\\
    &\mathbf S^i\mathbf l_i\leq\mathbf 0_N,\qquad \rho\geq0 .
    \label{eq:LP_feas1}
\end{align}
\blue{By construction, \eqref{eq:prop2} is feasible for the given polarity hypothesis whenever $\rho_i\geq\rho^l$.}

\vspace{0.05in}
\subsubsection{\textbf{Choosing $\rho_i$ for Previous Solution Feasibility}}
In subsequent iterations, we compute $\rho^l$ so that a previous solution $\mathbf{l}_i'$ remains feasible in \eqref{eq:prop2} \blue{(thus satisfies the current polarity constraint $\mathbf S^i\mathbf l_i'\leq\mathbf 0_N$)}. \blue{This supports the monotonicity argument in Section~\ref{subsec:convergence}.}
To accomplish this, we solve the following LP over $\rho$:
\begin{align}
\min_{\rho}\quad &\rho,\nonumber\\
\mathrm{s.t.}\quad
&\left\{
\begin{array}{l}
\mathbf e_i-\rho\mathbf 1_N
\leq\mathbf C\mathbf l_i'
\leq\mathbf e_i+\rho\mathbf 1_N,\\
\rho\geq0 .
\end{array}
\right.
\label{eq:LP_feas2}
\end{align}
\blue{Since $\mathbf l_i'$ is fixed, the solution is available in closed form:}
\begin{equation}
    \rho^l=\|\mathbf C\mathbf l_i'-\mathbf e_i\|_\infty .
\end{equation}
Consequently, any $\rho_i\geq\rho^l$ ensures that $\mathbf l_i'$ remains feasible. 

In our experiments, we use
\begin{equation}
    \rho_i=\rho^l+\delta_0,
\end{equation}
\blue{where the small margin $\delta_0>0$ avoids placing the previous solution exactly on the boundary of the covariance-fit constraints.} 

\subsection{Algorithm Termination}
\label{subsec:convergence}

Given our $\rho_i$ selection, \blue{we prove that our iterative algorithm terminates  in a finite number of steps (\textit{i.e.}, reach a solution matrix $\mathcal{L}^b$ where no LP solution $\mathbf{l}_i$ to optimization \eqref{eq:prop2} for column $i$ further decreases the optimization objective).}
We note first that given the overall objective $\| \mathcal{L} \|_1 = \sum_i \|\mathbf{l}_i\|_1$ is lower-bounded by zero, it is sufficient to prove \blue{algorithm termination} by showing that objective $\| \mathcal{L} \|_1$ is \textit{monotonically non-increasing} as the algorithm iterates. 

At the start of an iteration optimizing $\mathbf{l}_i$, the $i$th column / row of $\mathcal{L}$ has just been updated to $\mathbf{l}_i'$ during previous iteration optimizing $\mathbf{l}_{i-1}$; \textit{i.e.}, the $(i-1)$-th entry of $\mathbf{l}_i$ have been updated. 
For the current iteration, we choose $\rho$ large enough that $\mathbf{l}_i'$ remains in the feasible solution space (via LP \eqref{eq:LP_feas2}). With this choice, the optimal solution $\mathbf{l}^*_i$ obtained at iteration $t$ with the same polarity satisfies
\begin{align}
\|\mathbf{l}_i^* \|_1 \leq \|\mathbf{l}_i'\|_1 .
\end{align}
Node $i$ switches polarity only if the objective value obtained with opposite polarity is strictly smaller than $\|\mathbf{l}_i^*\|_1$. Hence, whether the polarity changes or not, $\|\mathcal{L}\|_1$ is monotonically non-increasing.
\blue{In practice, the algorithm terminates when a complete sweep over all columns produces no accepted objective decrease larger than a prescribed tolerance.}

\section{Fast Balanced Signed Graph Optimization}
\label{sec:opt}

\subsection{LP in Standard Form}
\label{subsubsec:LPform}

Towards efficient computation, we treat \eqref{eq:prop2} as a \textit{sparse linear programming} (SLP) problem\footnote{By SLP, we mean that the coefficient matrix $\mathbf{A}$ in the linear constraint $\mathbf{A} \mathbf{x} = \mathbf{b}$ in an LP standard formulation is sparse.}, efficiently solvable via a novel adaptation of ADMM that split variables for alternating optimizations until solution convergence \cite{wang17}. 

We first rewrite \eqref{eq:prop2} into LP standard form.
\blue{First, we exploit the sign constraint $\mathbf{S}^i \mathbf{l}_i \leq \mathbf{0}_N$ to linearize the $\ell_1$-norm objective as follows. 
\blue{Given the entry-wise constraint $S^i_{k,k} l_{i,k} \leq 0$, $\text{sign}(l_{i,k})$ must be $ -\text{sign}(S^i_{k,k})$. In other words, $l_{i,k} = - S^i_{k,k} m_k$ for $m_k \geq 0$. In vector form, we write  }
\begin{equation}
\mathbf{l}_i = -\mathbf{S}^i \mathbf{m}, \qquad \mathbf{m} \geq \mathbf{0}_N .
\label{eq:reduced_sub}
\end{equation}
The objective is now linear: $\|\mathbf{l}_i\|_1 = \mathbf{1}_N^\top \mathbf{m}$.}
Define also \textit{dependent variable} $\mathbf{r} \triangleq \mathbf{C} \mathbf{l}_i - \mathbf{e}_i = \mathbf{C} (-\mathbf{S}^i \mathbf{m}) - \mathbf{e}_i \in \mathbb{R}^{N}$.
This enables linear constraints $\pm r_j \leq \rho, \forall j$.

We now rewrite the set of linear constraints,
\blue{$-\mathbf{C} \mathbf{S}^i \mathbf{m} - \mathbf{r} = \mathbf{e}_i$},
$-\mathbf{r} \leq \blue{\rho_i} \mathbf{1}_N$,
$\mathbf{r} \leq \blue{\rho_i} \mathbf{1}_N$,
into the following aggregate equality constraint:

\vspace{-0.1in}
\begin{scriptsize}
\begin{align}
\underbrace{\left[ \begin{array}{c|c}
\begin{array}{cc}
\blue{-\mathbf{C}\mathbf{S}^i} & -\mathbf{I}_N \\
\mathbf{0}_{N,N} & \mathbf{I}_{N} \\
\mathbf{0}_{N,N} & -\mathbf{I}_{N}
\end{array} &
\begin{array}{c}
\mathbf{0}_{N,2N} \\
\mathbf{I}_{2N}
\end{array}
\end{array} \right]}_{\mathbf{A}}
\left[ \begin{array}{c}
\blue{\mathbf{m}} \\
\mathbf{r} \\
\mathbf{q}
\end{array}\right] &=
\underbrace{\left[ \begin{array}{c}
\mathbf{e}_i \\
\blue{\rho_i} \mathbf{1}_N \\
\blue{\rho_i} \mathbf{1}_N
\end{array} \right]}_{\mathbf{b}}
\label{eq:linearConstraint}
\end{align}
\end{scriptsize}\noindent
where $\mathbf{q}\geq\mathbf{0}_{2N} \in \mathbb{R}^{2N}$ is a non-negative \textit{slack variable}, so that we can convert inequality constraints to equality constraints. 
The elements of $\mathbf{S}^i$ are initialized as described in Section\;\ref{subsec:initialization}.
Note that aggregate matrix $\mathbf{A}$ is sparse except for \blue{$\mathbf{C}\mathbf{S}^i$} (to be further discussed in Section\;\ref{subsec:sparseCase}).

For notation convenience, define \blue{the stacked variable $\mathbf{y} \triangleq [\mathbf{m}; \mathbf{r}; \mathbf{q}] \in \mathbb{R}^{4N}$, \textit{cost vector} $\mathbf{c} \triangleq [\mathbf{1}_N; \mathbf{0}_{3N}] \in \mathbb{R}^{4N}$, and the selection matrix $\mathbf{E} \in \{0,1\}^{3N \times 4N}$ that extracts the non-negative sub-vector $\mathbf{z} \triangleq [\mathbf{m}; \mathbf{q}] = \mathbf{E}\mathbf{y}$}.
We can rewrite the objective as \blue{$\mathbf{c}^\top \mathbf{y}$}.
The resulting LP in standard form is
\begin{align}
\min_{\blue{\mathbf{y}}} \blue{\mathbf{c}^\top \mathbf{y}}, ~~~
\mbox{s.t.} ~ \blue{\mathbf{A} \mathbf{y} = \mathbf{b}, ~~
\mathbf{E}\mathbf{y} \geq \mathbf{0}_{3N}} .
\label{eq:LP}
\end{align}

\subsection{Fast Optimization via ADMM}
\label{subsec:optADMM}

We solve LP in \eqref{eq:LP} efficiently by tailoring an ADMM-based method \cite{wang17}.
We first define a convex but non-differentiable \textit{indicator} function:
\begin{align}
g(\blue{\tilde{\mathbf{z}}}) = \left\{ \begin{array}{ll}
0 & \mbox{if} ~ \blue{\tilde{z}_{j}} \geq 0, \forall j \\
\infty & \mbox{o.w.}
\end{array} \right. .
\end{align}

Next, to split the objective into two parts, we introduce \textit{auxiliary variable} $\tilde{\mathbf{z}} \in \mathbb{R}^{\blue{3N}}$ and rewrite optimization \eqref{eq:LP}:
\begin{align}
\min_{\blue{\mathbf{y}}, \blue{\tilde{\mathbf{z}}}} &~ \blue{\mathbf{c}^\top \mathbf{y}} + g(\blue{\tilde{\mathbf{z}}})
\label{eq:ADMM0} \\
\mbox{s.t.} & ~ \underbrace{\left[
\begin{array}{cc}
\mathbf{A} & \blue{\mathbf{0}_{3N,3N}} \\
\blue{\mathbf{E}} & \blue{-\mathbf{I}_{3N}} \end{array} \right]}_{\mathbf{B}}
\left[ \begin{array}{c}
\blue{\mathbf{y}} \\
\blue{\tilde{\mathbf{z}}}
\end{array} \right] =
\left[ \begin{array}{c}
\mathbf{b} \\
\blue{\mathbf{0}_{3N}}
\end{array} \right] .
\nonumber
\end{align}

We can now rewrite \eqref{eq:ADMM0} into an unconstrained version using the augmented Lagrangian method \cite{boyd11} as
\begin{align}
\min_{\blue{\mathbf{y}}, \blue{\tilde{\mathbf{z}}}} &~ \blue{\mathbf{c}^\top \mathbf{y}} + g(\tilde{\mathbf{z}}) + \boldsymbol{\mu}^\top \left( \mathbf{B} \left[ \begin{array}{c}
\blue{\mathbf{y}} \\
\blue{\tilde{\mathbf{z}}}
\end{array} \right] - \left[ \begin{array}{c}
\mathbf{b} \\
\blue{\mathbf{0}_{3N}}
\end{array} \right] \right)
\nonumber \\
& + \frac{\gamma}{2} \left\|
\mathbf{B} \left[ \begin{array}{c}
\blue{\mathbf{y}} \\
\blue{\tilde{\mathbf{z}}}
\end{array} \right] - \left[ \begin{array}{c}
\mathbf{b} \\
\blue{\mathbf{0}_{3N}}
\end{array} \right]
\right\|^2_2
\label{eq:ADMM}
\end{align}
where $\boldsymbol{\mu} \in \mathbb{R}^{\blue{6N}}$ is the
Lagrange multiplier vector \blue{corresponding to the $3N$ constraints in $\mathbf A\mathbf y=\mathbf b$ and the $3N$ constraints in $\mathbf E\mathbf y=\tilde{\mathbf z}$}, and $\gamma>0$ is a positive scalar parameter.

\vspace{0.1in}
\subsubsection{Optimizing Main Variables}
\label{subsub:main}

We minimize the unconstrained objective \eqref{eq:ADMM} alternately as follows.
At iteration $t$, when $\tilde{\mathbf{z}}^t$ is fixed, the optimization for \blue{$\mathbf{y}^{t+1}$} becomes
\begin{align}
\min_{\blue{\mathbf{y}}} &~ \blue{\mathbf{c}^\top \mathbf{y}} + \boldsymbol{\mu}^\top \left( \mathbf{B} \left[ \begin{array}{c}
\blue{\mathbf{y}} \\
\blue{\tilde{\mathbf{z}}^t}
\end{array} \right] - \left[ \begin{array}{c}
\mathbf{b} \\
\blue{\mathbf{0}_{3N}}
\end{array} \right] \right)
\nonumber \\
& + \frac{\gamma}{2} \left\|
\mathbf{B} \left[ \begin{array}{c}
\blue{\mathbf{y}} \\
\blue{\tilde{\mathbf{z}}^t}
\end{array} \right] - \left[ \begin{array}{c}
\mathbf{b} \\
\blue{\mathbf{0}_{3N}}
\end{array} \right]
\right\|^2_2 .
\label{eq:obj_main}
\end{align}
The solution to this convex and smooth quadratic optimization can be obtained by solving a linear system:
\begin{align}
& \gamma \underbrace{\left( \mathbf{A}^\top \mathbf{A} + \blue{\mathbf{E}^\top \mathbf{E}} \right)}_{\boldsymbol{\Psi}}
\blue{\mathbf{y}^{t+1}} =
\label{eq:linSys} \\
& \gamma \left( \blue{\mathbf{E}^\top \tilde{\mathbf{z}}^t} - \mathbf{A}^\top \mathbf{b} \right) - \mathbf{c}
-
\mathbf{A}^\top \boldsymbol{\mu}_1 - \blue{\mathbf{E}^\top \boldsymbol{\mu}_2}
\nonumber
\end{align}
where $\boldsymbol{\mu}_1 \in \mathbb{R}^{\blue{3N}}$ and $\boldsymbol{\mu}_2 \in \mathbb{R}^{\blue{3N}}$ are the multiplier sub-vectors such that $\boldsymbol{\mu} = [\boldsymbol{\mu}_1; \boldsymbol{\mu}_2]$.
See Appendix\;\ref{append:linSys} for a derivation.

\vspace{0.05in}
\noindent
\textbf{Complexity}:
Linear system \eqref{eq:linSys} can be solved efficiently using \textit{conjugate gradient} (CG) with complexity $\mathcal{O}(\text{nnz}(\boldsymbol{\Psi}) \sqrt{\kappa(\boldsymbol{\Psi})}\log(1/\epsilon))$, where $\text{nnz}(\boldsymbol{\Psi})$ is the number of non-zero entries in matrix $\boldsymbol{\Psi}$, $\kappa(\boldsymbol{\Psi}) = \frac{\lambda_{\max}(\boldsymbol{\Psi})}{\lambda_{\min}(\boldsymbol{\Psi})}$ is the \textit{condition number} of $\boldsymbol{\Psi}$,  $\lambda_{\max}(\boldsymbol{\Psi})$ and $\lambda_{\min}(\boldsymbol{\Psi})$ are the respective largest and smallest eigenvalues of $\boldsymbol{\Psi}$, and $\epsilon$ is the convergence threshold of the gradient search \cite{shewchuk94}.
$\kappa(\boldsymbol{\Psi})$ is reasonably small in practice.
\blue{In our implementation, \eqref{eq:linSys} is solved \textit{inexactly}: CG is warm-started from the previous ADMM iterate, and its tolerance follows a summable schedule $\epsilon_t = \epsilon_0/t^{p}$ with $p>1$, further tightened as the outer ADMM residuals decrease, which suffices for ADMM convergence under inexact subproblem minimization \cite{boyd11}.
Assuming $\epsilon_0$ is set reasonably, CG's execution time is $\mathcal{O}(\text{nnz}(\boldsymbol{\Psi})) = \mathcal{O}(N^2)$ due to dense $N$-by-$N$ sample covariance matrix $\mathbf{C}$ in $\mathbf{A}$ in \eqref{eq:linearConstraint}.}

\vspace{0.1in}
\subsubsection{Optimizing Auxiliary Variable}

Fixing computed \blue{$\mathbf{y}^{t+1}$, and writing $\mathbf{z}^{t+1} = \mathbf{E}\mathbf{y}^{t+1}$}, the optimization \eqref{eq:ADMM} for $\blue{\tilde{\mathbf{z}}^{t+1}}$ at iteration $t$ becomes
\begin{align}
\min_{\blue{\tilde{\mathbf{z}}}} & ~ g(\blue{\tilde{\mathbf{z}}}) + \boldsymbol{\mu}_2^\top \left( \left[ \blue{\mathbf{I}_{3N} ~~ -\mathbf{I}_{3N}} \right] \left[ \begin{array}{c}
\blue{\blue{\mathbf{z}}^{t+1}} \\
\blue{\tilde{\mathbf{z}}}
\end{array} \right] \right)
\nonumber \\
& + \frac{\gamma}{2} \left\|
\left[ \blue{\mathbf{I}_{3N} ~~ -\mathbf{I}_{3N}} \right] \left[ \begin{array}{c}
\blue{\blue{\mathbf{z}}^{t+1}} \\
\blue{\tilde{\mathbf{z}}}
\end{array} \right]
\right\|^2_2 .
\label{eq:auxVar}
\end{align}
The solution is a \textit{separable} thresholding operation.
Specifically, at iteration $t$, we compute $\tilde{\mathbf{z}}^{t+1}$ as
\begin{align}
\blue{\tilde{z}_j^{t+1}} = \left\{ \begin{array}{ll}
0 & \mbox{if} ~ \blue{\blue{z}_j^{t+1}} + \frac{1}{\gamma} \mu_{2,j}^t < 0 \\
\blue{z_j^{t+1}} + \frac{1}{\gamma} \mu_{2,j}^t & \mbox{o.w.}
\end{array} \right. .
\label{eq:threshold}
\end{align}
See Appendix\;\ref{append:threshold} for a derivation.

\vspace{0.05in}
\noindent
\textbf{Complexity}: 
Because \eqref{eq:threshold} is a simple addition executed entry-by-entry for $\mathcal{O}(N)$ entries, the complexity is $\mathcal{O}(N)$.

\vspace{0.1in}
\subsubsection{Updating Lagrange Multipliers}

After computing \blue{$\mathbf{y}^{t+1}$} and $\tilde{\mathbf{z}}^{t+1}$, Lagrange multiplier vector $\boldsymbol{\mu}^{t+1}$ can be updated at iteration $t$ in the usual manner in ADMM \cite{boyd11}:
\begin{align}
\boldsymbol{\mu}^{t+1} = \boldsymbol{\mu}^t + \gamma \left( \mathbf{B} \left[ \begin{array}{c}
\blue{\mathbf{y}^{t+1}} \\
\blue{\tilde{\mathbf{z}}^{t+1}}
\end{array} \right] - \left[ \begin{array}{c}
\mathbf{b} \\
\blue{\mathbf{0}_{3N}}
\end{array} \right] \right) .
\label{eq:multiplierUpdate}
\end{align}
\textbf{Complexity}:
\eqref{eq:multiplierUpdate} requires one matrix-vector multiplication and simple vector additions, where $\text{nnz}(\mathbf{B}) = \mathcal{O}(N^2)$, again due to dense $N$-by-$N$ matrix \blue{$\mathbf{C}\mathbf{S}^i$} in $\mathbf{A}$ in \eqref{eq:linearConstraint}.
Thus, the complexity is $\mathcal{O}(N^2)$. 

\subsection{Optimizing the $K \ll N$  Case}
\label{subsec:sparseCase}

In the special case when $K \ll N$, we can further sparsify the aggregate matrix $\mathbf{A}$ in the LP formulation \eqref{eq:LP}, so that optimizing the main variables \blue{$\mathbf{y}^{t+1}$} in \eqref{eq:linSys} via CG and updating multipliers $\boldsymbol{\mu}^{t+1}$ in \eqref{eq:multiplierUpdate} both executes in $\mathcal{O}(N)$. 
The basic idea is to write $\mathbf{X}$ and $\mathbf{X}^\top$ separately in the aggregate constraint, instead of pre-computing dense sample covariance matrix \blue{$\mathbf{C} = \tfrac{1}{K-1} \mathbf{X} \mathbf{X}^\top$} with $N^2$ nonzero entries.

Recall that $\mathbf{r} \triangleq \mathbf{C} \mathbf{l}_i - \mathbf{e}_i$ in Section\;\ref{subsubsec:LPform}.
Instead, we replace it with two equivalent linear constraints:
\begin{align}
\mathbf{r} &= \tfrac{1}{K-1} \mathbf{X} \boldsymbol{\phi} - \mathbf{e}_i
\nonumber \\
\boldsymbol{\phi} &= \mathbf{X}^\top \mathbf{l}_i \blue{= -\mathbf{X}^\top \mathbf{S}^i \mathbf{m}}
\end{align}
where $\boldsymbol{\phi} \in \mathbb{R}^K$ is an \textit{intermediate variable}.
Replacing the first constraint in the aggregate linear constraint \eqref{eq:linearConstraint} with these two, we get

\vspace{-0.1in}
\begin{scriptsize}
\begin{align}
\underbrace{\left[ \begin{array}{c|c}
\begin{array}{ccc}
\blue{\mathbf{0}_{N,N}} & -\mathbf{I}_N & \tfrac{1}{K-1}\mathbf{X} \\
\blue{-\mathbf{X}^\top \mathbf{S}^i} & \mathbf{0} & -\mathbf{I}_K \\
\blue{\mathbf{0}_{N,N}} & \mathbf{I}_{N} & \mathbf{0}_{N,K} \\
\blue{\mathbf{0}_{N,N}} & -\mathbf{I}_{N} & \mathbf{0}_{N,K}
\end{array} &
\!\!\!\!\!
\begin{array}{c}
\mathbf{0}_{N,2N} \\
\mathbf{0}_{K,2N} \\
\blue{\mathbf{I}_{2N}}
\end{array}
\!\!\!\!\!
\end{array} \!\! \right]}_{\mathbf{A}}
\!\!
\left[ \!\! \begin{array}{c}
\blue{\mathbf{m}} \\
\mathbf{r} \\
\boldsymbol{\phi} \\
\mathbf{q}
\end{array} \!\! \right]
\!\! &= \!\!
\underbrace{\left[ \!\! \begin{array}{c}
\mathbf{e}_i \\
\mathbf{0}_K \\
\blue{\rho_i \mathbf{1}_N} \\
\blue{\rho_i \mathbf{1}_N}
\end{array} \!\! \right]}_{\mathbf{b}}
\nonumber \\
& \mathbf{q} \geq \blue{\mathbf{0}_{2N}} .
\label{eq:linearConstraint2}
\end{align}
\end{scriptsize}\noindent
The number of nonzero entries in aggregate matrix $\mathbf{A}$ is now $\mathcal{O}(N)$ for $K \ll N$.
Subsequently, matrix $\mathbf{B}$ (for \blue{$\mathbf{y} = [\mathbf{m}; \mathbf{r}; \boldsymbol{\phi}; \mathbf{q}]$, with $\mathbf{E}$ extracting $[\mathbf{m}; \mathbf{q}]$}) also has $\mathcal{O}(N)$ nonzero entries.
Hence, CG used to solve the linear system \eqref{eq:linSys} is also $\mathcal{O}(N)$.
Further, updating multiplier $\boldsymbol{\mu}^{t+1}$ in \eqref{eq:multiplierUpdate} given $\mathcal{O}(N)$ nonzero entries in matrix $\mathbf{B}$ is also $\mathcal{O}(N)$.

\subsection{Choosing CLIME Parameter $\rho_i$}
\label{subsec:select_rho}
 
Similar to Section~\ref{subsec:rho}, the minimum feasible parameter $\rho^l$ in the $K\ll N$ case can be computed by solving the LP \blue{(in the reduced variable $\mathbf{m}$ of \eqref{eq:reduced_sub})}:
\begin{align}
\min_{\blue{\mathbf{m}},\rho, \boldsymbol{\phi}} ~\rho,
~~~\mbox{s.t.}~~
\left\{ \begin{array}{l}
\mathbf{e}_i - \mathbf{1} \rho \leq \frac{1}{K-1}\mathbf{X} \boldsymbol{\phi} \leq \mathbf{e}_i + \mathbf{1} \rho
\\
\boldsymbol{\phi} = \blue{-\mathbf{X}^\mathsf{T} \mathbf{S}^i \mathbf{m}}
\\
\blue{\mathbf{m} \geq \mathbf{0}_N}
\\
\rho \geq 0
\end{array} \right. .
\end{align}

If a previous solution $\mathbf{l}_i'$ is available, then we solve the following LP instead:
\begin{align}
\min_{\rho,\boldsymbol{\phi}} ~\rho, 
~~~\mbox{s.t.}~~
\left\{ \begin{array}{l}
\mathbf{e}_i - \mathbf{1} \rho \leq \frac{1}{K-1}\mathbf{X} \boldsymbol{\phi} \leq \mathbf{e}_i + \mathbf{1} \rho 
\\
\boldsymbol{\phi} = \mathbf{X}^\mathsf{T} \mathbf{l}_i'
\\
\rho \geq 0
\end{array} \right. 
\end{align}
assuming $\mathbf{S}^i \mathbf{l}_i' \leq \mathbf{0}_N$\blue{; since $\mathbf{l}_i'$ is fixed, its solution is available in closed form, $\rho^l = \| \frac{1}{K-1}\mathbf{X}\boldsymbol{\phi} - \mathbf{e}_i \|_\infty$}. 

\blue{We then set $\rho_i = \rho^l + \delta_0$ for a small margin $\delta_0 > 0$: at $\rho_i = \rho^l$ the feasible set of \eqref{eq:prop2} has an empty interior, which slows the ADMM iterations considerably, while the margin keeps the LP strictly feasible.}

\subsection{\blue{Two-stage Solver for Degenerate LPs}}

\blue{When $K<N$, $\mathbf C=\frac{1}{K-1}\mathbf X\mathbf X^\top$
is rank-deficient and $\dim\mathcal N(\mathbf C)\geq N-K$. Hence,
for any $\mathbf u\in\mathcal N(\mathbf C)$,
\begin{equation}
    \|\mathbf C(\mathbf l_i+\mathbf u)-\mathbf e_i\|_\infty
    =
    \|\mathbf C\mathbf l_i-\mathbf e_i\|_\infty .
\end{equation}
Thus, nullspace perturbations leave the covariance fit unchanged and
can produce multiple optimal solutions, provided that they also satisfy
the sign constraint in \eqref{eq:prop2}.}

\blue{We optionally resolve this ambiguity using two LPs. Stage~1
solves \eqref{eq:prop2} and returns $\mathbf m^{(1)}$ with optimal
value
$z_i^\star=\mathbf 1_N^\top\mathbf m^{(1)}
=\|\mathbf l_i^{(1)}\|_1$.
Stage~2 is warm-started from $\mathbf m^{(1)}$ and solves
\begin{align}
    \max_{\mathbf m}\quad
    &m_i-\sum_{j\neq i}m_j \nonumber\\
    \mathrm{s.t.}\quad
    &\|-\mathbf C\mathbf S^i\mathbf m-\mathbf e_i\|_\infty
      \leq\rho_i,\nonumber\\
    &\mathbf m\geq\mathbf 0_N,\qquad
      \mathbf 1_N^\top\mathbf m\leq z_i^\star .
    \label{eq:two_stage_degenerate}
\end{align}
The last constraint keeps Stage~2 within the Stage-1 optimal set.
Its objective maximizes the $i$-th Gershgorin disc left-end,
$l_{i,i}-\sum_{j\neq i}|l_{i,j}|$, and therefore favors diagonal
dominance. 
This second LP selects a numerically stable solution to resolve solution ambiguity.}

\subsection{Algorithm Complexity Analysis}
\label{subsec:complexity}

We analyze the complexity of our algorithm.
For each column $\mathbf{l}_i$ in balanced signed graph Laplacian $\mathcal{L}^b$, we execute the ADMM-based algorithm in Section\;\ref{subsec:optADMM} twice to minimize objective \eqref{eq:ADMM}, where the complexities to compute the main variables, auxiliary variable, and update Lagrange multipliers are $\mathcal{O}(N^2)$, $\mathcal{O}(N)$ and $\mathcal{O}(N^2)$, respectively, if the dense covariance matrix $\mathbf{C}$ is pre-computed.
The number of iterations in ADMM till convergence is a constant that is not a function of $N$.
Given $N$ columns in matrix $\mathcal{L}^b$, the algorithm complexity is $\mathcal{O}(N^3)$. 

For the $K\ll N$ case, using aggregate linear constraint \eqref{eq:linearConstraint2} instead, where matrix $\mathbf{A}$ has $\mathcal{O}(N)$ non-zero entries, the complexities to compute the main and auxiliary variables and Lagrange multipliers in one ADMM iteration are all $\mathcal{O}(N)$.
Thus, the algorithm complexity is $\mathcal{O}(N^2)$.

\section{Experiments}
\label{sec:results}

In this section, we present experimental results for a comprehensive evaluation of the proposed balanced graph learning algorithm. 
We first compare the accuracy of balanced graph learning schemes using a synthetic dataset. 
We test the performance on both noisy and noiseless settings. Then, we test competing schemes on real-world datasets and evaluate their performance in graph signal restoration tasks.

\subsection{Synthetic Graph Recovery under Observation Noise}
\label{sec:synthetic_noise}

\begin{table*}[tb]
\centering
\caption{\blue{Sign-aware edge recovery and precision-matrix error under varying observation noise (mean over 10 trials).}}
\label{table:synth1}
\resizebox{\textwidth}{!}{%
\begin{tabular}{@{}l ccccc ccccc ccccc@{}}
\toprule
& \multicolumn{5}{c}{Macro-F1 $\uparrow$}
& \multicolumn{5}{c}{POL $\uparrow$}
& \multicolumn{5}{c}{RE $\downarrow$} \\
\cmidrule(lr){2-6}\cmidrule(lr){7-11}\cmidrule(l){12-16}
Method & $\sigma{=}0$ & $0.1$ & $0.2$ & $0.3$ & $0.4$
       & $\sigma{=}0$ & $0.1$ & $0.2$ & $0.3$ & $0.4$
       & $\sigma{=}0$ & $0.1$ & $0.2$ & $0.3$ & $0.4$ \\
\midrule
\multicolumn{16}{@{}c}{\emph{Sample-rich regime ($N{=}50$, $K{=}2000$)}} \\
\midrule
Proposed        & $\mathbf{.840}$ & $\mathbf{.834}$ & $\mathbf{.828}$ & $\mathbf{.827}$ & $\mathbf{.816}$
                & $\underline{.980}$ & $\underline{.978}$ & $.978$ & $.976$ & $.976$
                & $\mathbf{.162}$ & $\mathbf{.169}$ & $\mathbf{.191}$ & $\mathbf{.228}$ & $\mathbf{.277}$ \\
BsigGL          & $\underline{.733}$ & $\underline{.675}$ & $\underline{.700}$ & $\underline{.673}$ & $.639$
                & $\underline{.980}$ & $\mathbf{.982}$ & $\mathbf{.984}$ & $\mathbf{.986}$ & $\mathbf{.988}$
                & $.454$ & $.457$ & $.458$ & $.458$ & $.459$ \\
CLIME (mincut)  & $.595$ & $.595$ & $.626$ & $.622$ & $.580$
                & $.566$ & $.548$ & $.558$ & $.568$ & $.544$
                & $.284$ & $.288$ & $.302$ & $.329$ & $.364$ \\
CLIME (maxcut)  & $.618$ & $.632$ & $.619$ & $.631$ & $.603$
                & $.550$ & $.572$ & $.560$ & $.532$ & $.528$
                & $.281$ & $.288$ & $.304$ & $.330$ & $.364$ \\
CLIME (greedy)  & $.647$ & $.652$ & $.648$ & $.646$ & $.646$
                & $.976$ & $.976$ & $.976$ & $.880$ & $.786$
                & $.271$ & $.276$ & $.294$ & $.323$ & $.358$ \\
GLasso (mincut) & $.268$ & $.308$ & $.284$ & $.247$ & $.288$
                & $.584$ & $.594$ & $.596$ & $.588$ & $.582$
                & $.239$ & $.242$ & $.258$ & $.290$ & $.328$ \\
GLasso (maxcut) & $.224$ & $.276$ & $.294$ & $.282$ & $.197$
                & $.592$ & $.566$ & $.562$ & $.574$ & $.556$
                & $.233$ & $.239$ & $.255$ & $.285$ & $.323$ \\
GLasso (greedy) & $.648$ & $.650$ & $.650$ & $.651$ & $\underline{.651}$
                & $\mathbf{.982}$ & $.888$ & $\underline{.980}$ & $\underline{.980}$ & $\underline{.982}$
                & $\underline{.213}$ & $\underline{.220}$ & $\underline{.242}$ & $\underline{.277}$ & $\underline{.315}$ \\
\midrule
\multicolumn{16}{@{}c}{\emph{Sample-starved regime ($N{=}100$, $K{=}50$)}} \\
\midrule
Proposed        & $\underline{.110}$ & $\mathbf{.109}$ & $\mathbf{.104}$ & $\mathbf{.105}$ & $\mathbf{.098}$
                & $\mathbf{.785}$ & $\mathbf{.773}$ & $\mathbf{.771}$ & $\mathbf{.767}$ & $\mathbf{.758}$
                & $.447$ & $.453$ & $.462$ & $.472$ & $.485$ \\
BsigGL          & $\mathbf{.126}$ & $\underline{.101}$ & $\underline{.093}$ & $\underline{.094}$ & $\underline{.096}$
                & $\underline{.756}$ & $\underline{.764}$ & $\underline{.748}$ & $\underline{.722}$ & $\underline{.715}$
                & $.523$ & $.524$ & $.524$ & $.523$ & $.526$ \\
CLIME (mincut)  & $.015$ & $.014$ & $.003$ & $.009$ & $.000$
                & $.530$ & $.544$ & $.539$ & $.545$ & $.534$
                & $.406$ & $.410$ & $.408$ & $\underline{.402}$ & $\mathbf{.387}$ \\
CLIME (maxcut)  & $.031$ & $.023$ & $.004$ & $.008$ & $.002$
                & $.547$ & $.534$ & $.541$ & $.536$ & $.557$
                & $.406$ & $\underline{.409}$ & $\underline{.400}$ & $\underline{.402}$ & $\mathbf{.387}$ \\
CLIME (greedy)  & $.014$ & $.011$ & $.004$ & $.003$ & $.006$
                & $.541$ & $.574$ & $.538$ & $.547$ & $.549$
                & $.406$ & $\underline{.409}$ & $.408$ & $\underline{.402}$ & $\mathbf{.387}$ \\
GLasso (mincut) & $.008$ & $.000$ & $.002$ & $.002$ & $.002$
                & $.545$ & $.554$ & $.518$ & $.545$ & $.549$
                & $\underline{.376}$ & $\mathbf{.378}$ & $\mathbf{.384}$ & $\mathbf{.394}$ & $\underline{.408}$ \\
GLasso (maxcut) & $.002$ & $.001$ & $.000$ & $.002$ & $.001$
                & $.542$ & $.529$ & $.546$ & $.526$ & $.545$
                & $\mathbf{.375}$ & $\mathbf{.378}$ & $\mathbf{.384}$ & $\mathbf{.394}$ & $\underline{.408}$ \\
GLasso (greedy) & $.003$ & $.003$ & $.002$ & $.002$ & $.002$
                & $.540$ & $.557$ & $.559$ & $.550$ & $.534$
                & $\underline{.376}$ & $\mathbf{.378}$ & $\mathbf{.384}$ & $\mathbf{.394}$ & $\underline{.408}$ \\
\bottomrule
\end{tabular}}
\vspace{-10pt}
\end{table*}

\blue{We evaluate estimation accuracy using three metrics: sign-aware macro-F1, polarity accuracy (POL), and relative precision-matrix error (RE).}
\blue{Macro-F1 measures whether both the support and the sign of each edge are recovered correctly.
Let $\mathcal{E}_{+}^{\ast}$ and $\mathcal{E}_{-}^{\ast}$ be the sets of positive and negative edges in the ground-truth graph.
Similarly, let $\widehat{\mathcal{E}}_{+}$ and $\widehat{\mathcal{E}}_{-}$ be the sets of positive and negative edges in the estimated graph.
For each sign class $s\in\{+,-\}$, precision and recall are defined as
\begin{align}
    \mathrm{Prec}_{s}
    =
    \frac{|\widehat{\mathcal{E}}_{s}\cap\mathcal{E}_{s}^{\ast}|}
    {|\widehat{\mathcal{E}}_{s}|},
    \qquad
    \mathrm{Rec}_{s}
    =
    \frac{|\widehat{\mathcal{E}}_{s}\cap\mathcal{E}_{s}^{\ast}|}
    {|\mathcal{E}_{s}^{\ast}|}.
\end{align}
The F-measure for class $s$ is then
\begin{align}
    \mathrm{F1}_{s}
    =
    \frac{2\,\mathrm{Prec}_{s}\mathrm{Rec}_{s}}
    {\mathrm{Prec}_{s}+\mathrm{Rec}_{s}}.
\end{align}
The sign-aware macro-F1 is defined as
\begin{align}
    \text{Macro-F1}
    =
    \frac{1}{2}\left(\mathrm{F1}_{+}+\mathrm{F1}_{-}\right).
\end{align}
Thus, an edge estimated with the wrong sign is counted as an error for both sign classes.}

\blue{Polarity accuracy measures whether the node polarities are recovered correctly.
Since the polarity vector is identifiable only up to a global sign flip, we define
\begin{align}
    \mathrm{POL}
    =
    \max\left\{
    \frac{1}{N}\sum_{i=1}^{N}\mathbf{1}\{\widehat{\beta}_{i}=\beta_i^{\ast}\},
    \frac{1}{N}\sum_{i=1}^{N}\mathbf{1}\{-\widehat{\beta}_{i}=\beta_i^{\ast}\}
    \right\}.
\end{align}}

\blue{We also evaluate the accuracy of the estimated precision matrix with the relative error (RE), which is defined as}
\begin{equation}
	\text{RE} = \frac{\| \widehat{\mathcal{L}} - \mathcal{L}^{\ast} \|_{F} }{\| \mathcal{L}^{\ast} \|_{F} },
\end{equation}
where $\widehat{\mathcal{L}}$ is the estimate and $\mathcal{L}^\ast$ is the ground truth \blue{precision matrix}.

\subsubsection{Setting} 
\blue{The ground-truth graph $\mathcal{G}$ is an Erd\H{o}s--R\'{e}nyi graph \cite{Erdos1959pmd} on $N$ nodes with expected degree $4$. 
Denote by $\mathbf{A}$ the binary adjacency matrix. Observations are drawn as $\mathbf{x}\sim\mathcal{N}(\mathbf{0},\mathbf{S}\mathbf{Q}^{-1}\mathbf{S})$, where
\begin{equation}
\label{eq:synth_prec}
    \mathbf{Q}=\mathbf{D}^{1/2}
    \left(\mathbf{I}_N-\alpha
    \frac{\mathbf{A}}{\|\mathbf{A}\|_2}\right)
    \mathbf{D}^{1/2},\qquad \alpha=0.8.
\end{equation}
and $\mathbf{S}=\operatorname{diag}(\boldsymbol{\beta})$. The elements of $\boldsymbol{\beta}$ are polarities $\beta_i\in\{-1,1\}$ drawn independently with equal probability.
The computation inside the brackets sets the conditional dependence structure. 
Its off-diagonal support coincides with the edge set of $\mathcal{G}$. 
Since $\|\mathbf{A}/\|\mathbf{A}\|_2\|_2=1$ and $\alpha<1$, its eigenvalues lie in $[1-\alpha,1+\alpha]$, so $\mathbf{Q}\succ\mathbf{0}$ for every realization. 
Then we apply a node-wise scaling, with $D_{ii}$ drawn i.i.d. from $\mathcal{U}[0.4,2.5]$. 
This leaves the support of $\mathbf{Q}$ unchanged, while making the conditional variances $1/Q_{ii}=1/D_{ii}$ node-dependent, so that the entries of $\mathbf{Q}$ are spread over a range of magnitudes rather than being uniform. 
Since $Q_{ij}\neq 0$ if and only if $(i,j)\in\mathcal{E}$, recovering $\mathcal{G}$ is equivalent to recovering the off-diagonal support of $\mathbf{Q}$.}

\blue{We consider a sample-rich regime with $N=50$, $K=2000$, and a sample-starved regime with $N=100$, $K=50$.
The observations are corrupted by additive white Gaussian noise with standard deviation $\sigma\in\{0,0.1,0.2,0.3,0.4\}$, and all results are averaged over 10 independent trials.}

\blue{For the proposed method, the graph Laplacian is computed with the ADMM solver of Section\;\ref{sec:opt} (residual tolerance $10^{-3}$, at most $1000$ iterations per column, adaptive penalty, feasibility margin $\delta_0 = 10^{-4}$), with node polarities initialized by the Lanczos-based initializer in Section \ref{subsec:initialization}.}

The performance of our algorithm is compared against several two-step graph balancing approaches, i.e., a precision matrix estimation followed by a graph balancing step. 
We employed CLIME \cite{Cai_aconstrained_2011} or GLASSO \cite{mazumder_graphical_2012} as the baseline precision matrix estimation methods. 
The three variants of graph balancing methods were;
\begin{description}[labelwidth=12em]
    \item[\textit{MinCut Balancing (Min)} \cite{yokota_signed_2023}:] Nodes were polarized based on the min-cut of positive edges, and inconsistent edges were removed. 
    \item[\textit{MaxCut Balancing (Max)} \cite{yokota_signed_2023}:] Nodes were polarized based on the max-cut of negative edges, and inconsistent edges were removed. 
    \item[\textit{Greedy Balancing (Greed)} \cite{Dinesh2025}:] A polarized set was initialized by polarizing a random node to $1$, then nodes connected to the set were greedily polarized one by one, keeping the edge with the largest weight magnitude and removing the rest in each polarization.
\end{description}
As a result, alternative methods have six configurations.
\blue{Each balanced baseline graph is then calibrated by the validation procedure described above.
We also compared against BsigGL \cite{matz23}, a recent balanced graph learning method that learns edge signs and weight magnitude using a semi-definite programming and quadratic optimization.
BsigGL employs squared Frobenius-norm regularization instead of an explicit $\ell_1$ sparsity-promoting term and constrains the nonnegative weight matrix via fixed total-degree normalization. It has pre-defined bounds on edge weights and node degrees---parameters that are difficult to select in the absence additional domain knowledge. Thus, its graph density depends on the regularization parameter $\mu$ and the chosen feasible set $\mathcal{W}$. In our experiments, its estimates are dense. In contrast, our CLIME-based formulation directly promotes sparsity through $\ell_1$-minimization and uses a \textit{single} margin, $\delta_0$, to set all column-wise tolerances, $\rho_i=\rho_l+\delta_0$, ensuring feasibility of every sign-constrained column problem.
}

\blue{For a fair comparison, we post-process every output identically, in the following manner.
First, off-diagonal entries below a threshold $\tau$ (measured relative to the diagonal) are set to zero, yielding the sparsified estimate $\widehat{\mathbf{L}}_{\tau}$.
Second, the result is rescaled and regularized as $c\,\widehat{\mathcal{L}}_{\tau}+\rho\mathbf{I}$, which places all methods on a common scale and guarantees positive definiteness.
The parameters $(\tau,c,\rho)$ are selected by minimizing the Gaussian negative log-likelihood of $K_{\mathrm{val}}=200$ held-out noisy observations under the observation model $(c\,\widehat{\mathbf{L}}_{\tau}+\rho\mathbf{I})^{-1}+\sigma^{2}\mathbf{I}$.}

\subsubsection{Results}
\blue{
Table~\ref{table:synth1} summarizes the results.
In the sample-rich regime, the proposed method attains the best macro-F1 \emph{and} the best RE at every noise level, with mild degradation at $\sigma{=}0.4$, and its margin over the baselines widens with noise.
Among the two-step baselines, only the greedy-balanced variants remain competitive in macro-F1: mincut and maxcut balancing frequently mis-assign node polarities, and removing the resulting inconsistent edges also degrades their macro-F1, most severely for GLASSO.
The proposed method, BsigGL, and the greedy variants recover polarities almost perfectly in this regime, although the greedy variants become unstable at high noise.
BsigGL's RE saturates around $0.46$ regardless of noise due to the weight bias introduced by its normalization constraint.
}

\blue{
In the sample-starved regime, signed-edge recovery is difficult for all methods. The nonzero partial correlations have magnitude at most $\alpha/\|\mathbf A\|_2\approx0.15$, which is comparable to the sampling variability expected from \(K=50\) observations. Accordingly, macro-F1 remains below \(0.13\) for all methods.
Under the validation-based calibration, the CLIME/GLASSO estimates are pruned to (near-)empty graphs.
Although their near-diagonal estimates consequently yield the smallest RE, the estimate carries almost no off-diagonal structure. 
The proposed method and BsigGL still return meaningful balanced structure; the proposed method attains the best macro-F1 for $\sigma\ge 0.1$, the best polarity accuracy at all noise levels, and a lower RE than BsigGL throughout, showing that the dominant balanced structure remains recoverable even when individual edges are not.
}

\subsection{\blue{Frustration Study}}
\label{sec:synthetic_frustration}

\begin{table}[tb]
\centering
\caption{\blue{Proposed method under a nearly balanced ground truth with a fraction $\eta$ of frustrated edges.}}
\label{table:frustration}
\begin{tabular}{@{}l ccccc@{}}
\toprule
& \multicolumn{5}{c}{Frustrated-edge fraction} \\
\cmidrule(l){2-6}
& $\eta{=}0$ & $0.025$ & $0.05$ & $0.10$ & $0.15$ \\
\midrule
Macro-F1                        & $.840$ & $.828$ & $.803$ & $.753$ & $.721$ \\
RE                              & $.162$ & $.186$ & $.206$ & $.241$ & $.261$ \\
POL                             & $.980$ & $.966$ & $.948$ & $.888$ & $.828$\\
\bottomrule
\end{tabular}
\vspace{-10pt}
\end{table}

\blue{
As discussed in Section \ref{sec:prop}, real-world signed networks tend towards \textit{near-balanced} structures rather than \textit{perfectly-balanced} \cite{yu2024spreading,peiyao2025robust}.
Here, we examine the robustness of the proposed method when the data-generating precision matrix contain a few \emph{frustrated} edges, \ie, edges with signs inconsistent with the polarity.}

\blue{Let $\mathcal{L}_0^\ast$ denote the balanced precision matrix from the sample-rich, noiseless setting of Section \ref{sec:synthetic_noise}. For each frustration level $\eta$, we flip a nested fraction $\eta$ of its edge signs by replacing $\mathbf{A}$ in \eqref{eq:synth_prec} with $\mathbf{A}_\eta=\mathbf{F}_\eta\odot\mathbf{A}$, where $\mathbf{F}_\eta=\{-1,1\}^{N\times N}$ is symmetric binary matrix. Since $|\mathbf{A}_\eta|=\mathbf{A}$ and $\|\mathbf{A}_\eta\|_2\leq\|\mathbf{A}\|_2$, the resulting $\mathcal{L}_\eta^\ast$ remains positive definite for $\alpha<1$. We draw $\mathbf{x}^k\sim\mathcal{N}(\mathbf{0}_N,(\mathcal{L}_\eta^\ast)^{-1})$ with all other settings unchanged.}

\blue{Table~\ref{table:frustration} summarizes the results.
The macro-F1 of the proposed method consistently recovers approximately $84$--$85\%$ of the non-frustrated edges, indicating consistency with the original balanced structure. The RE also shows gradual degradation as the data-generating graph departs from exact structural balance. Overall, the results suggests that the proposed method is capable of recovering the major balanced component when the underlying signed graph is nearly balanced.
}

\subsection{Runtime and Memory Comparison}
\label{subsec:runtime}

\begin{table}[tb]
  \centering
  \caption{\blue{Median per-column \texttt{linprog} time, per-CG-iteration time of the ADMM column solvers.}}
  \label{tab:runtime_N}
  \begin{tabular}{rrrr|rr}
    \toprule
    & \multirow{2}{*}{LP [s/col.]} & \multicolumn{2}{c}{ADMM [ms/CG iter.]} &
      \multicolumn{2}{c}{Memory [MB]} \\
    \cmidrule(lr){3-4}\cmidrule(lr){5-6}
    $N$ & & dense & mat--vec & dense & mat--vec \\
    \midrule
     200 &   0.049 &  0.065 & 0.049 & 1.35 & 0.35 \\
     400 &   0.125 &  0.171 & 0.093 & 3.94 & 0.70 \\
     800 &   0.647 &  0.517 & 0.193 & 13.0 & 1.40 \\
    1600 &   4.70  &  2.10  & 0.440 & 46.4 & 2.79 \\
    3200 &  33.5   &  7.66  & 0.866 & 175  & 5.58 \\
    6400 & 227     &  --    & 1.70  & --   & 11.2 \\
    \bottomrule
  \end{tabular}
\end{table}

\begin{figure}[tb]
  \centering
  \subfigure[Fixed $K=50$]{\includegraphics[width=0.48\linewidth]{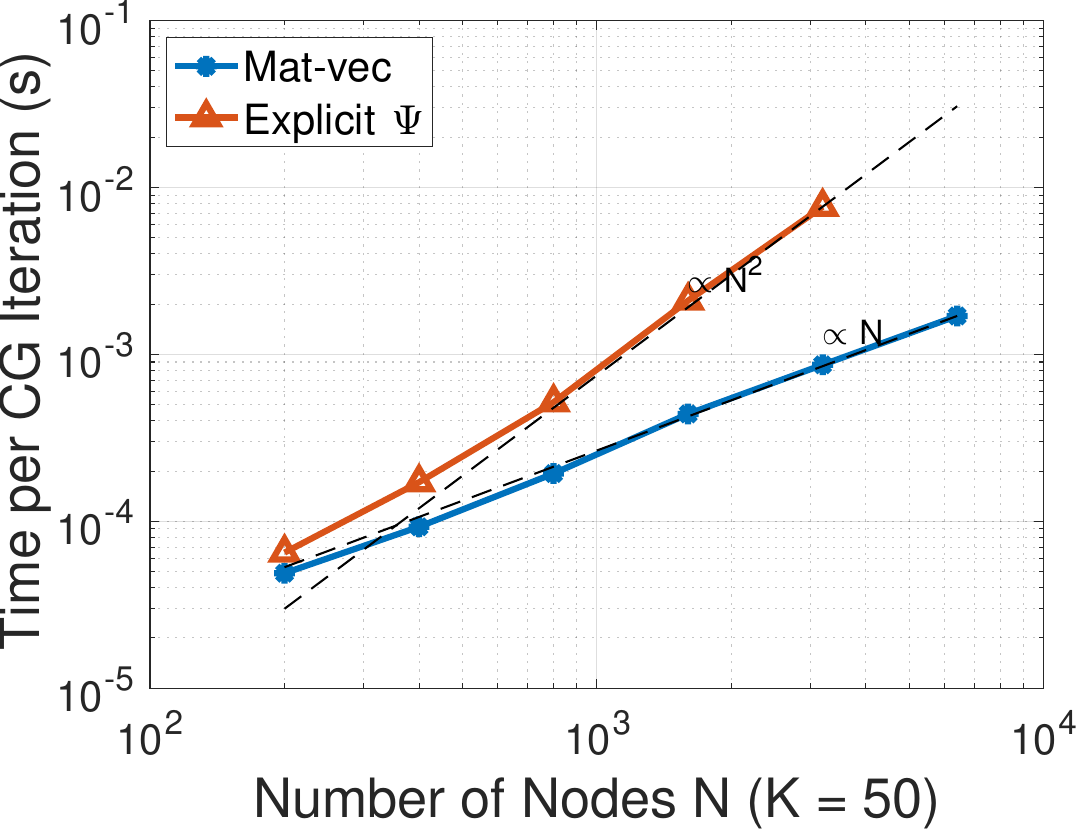}}
  \subfigure[Fixed $N=2000$]{\includegraphics[width=0.48\linewidth]{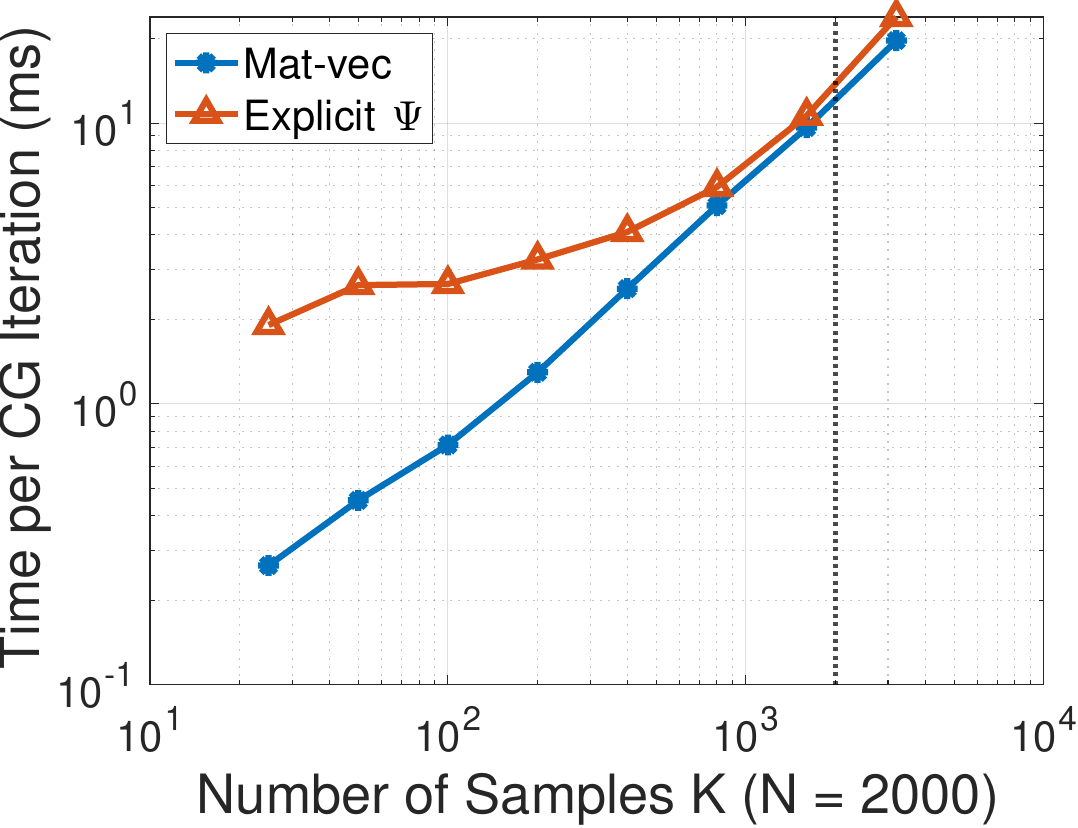}}
  \caption{\blue{Per-CG-iteration runtime of the ADMM column solver. Left: sweep over $N$ at fixed $K=50$. Right: sweep over $K$ at fixed $N=2000$. The matrix--vector-product-based (mat--vec) implementation scales as $O(NK)$, while explicitly forming $\boldsymbol{\Psi}$ in \eqref{eq:linSys} costs $O(N^2)$ per CG iteration for $K \ll N$.}}
  \label{fig:runtime}
\end{figure}

\blue{
We next evaluate the computational cost of the proposed ADMM solver.
The goal is to verify the per-iteration complexity of the proposed implementation in the $K\ll N$ regime.
We compare three solvers for the same reduced column-wise LP \eqref{eq:LP}, with the same $\rho_i$ from \eqref{eq:LP_feas1} per column: the proposed ADMM solver, which applies the sparse constraint matrix only through matrix--vector products (``mat--vec''); the identical ADMM solver with the matrix $\boldsymbol{\Psi}$ in \eqref{eq:linSys} formed explicitly (its $\mathbf{m}$-block contains the dense $N \times N$ matrix $(\mathbf{S}^i\mathbf{X})(\mathbf{S}^i\mathbf{X})^\top$); and MATLAB \texttt{linprog}.
The two ADMM variants use the same update equations and produce matching iterates.
Thus, their difference comes only from whether $\boldsymbol{\Psi}$ is formed, and the relevant unit is the cost of one inner CG iteration.
Instances follow the synthetic model of Section \ref{sec:synthetic_noise}.
}

\blue{
We fix $K=50$ and vary $N$.
Table~\ref{tab:runtime_N} summarizes the per-CG-iteration runtime and memory usage, and Fig.~\ref{fig:runtime} plots the scaling.
The proposed mat--vec solver scales roughly linearly with $N$, matching the theoretical $O(NK)$ complexity, which is $O(N)$ for fixed $K$.
The explicit-$\boldsymbol{\Psi}$ variant has slope approaching the expected $O(N^2)$, and the gap widens with $N$.
Owing to its $O(N^2)$ cost and memory, this variant is run only up to $N=3200$.
The per-column \texttt{linprog} time grows with slope $2.5$. The memory usage shows the same trend.}

\blue{We also vary $K$ at fixed $N=2000$ (Fig.~\ref{fig:runtime}, right).
The mat--vec cost per CG iteration grows approximately linearly with $K$, as expected from $O(NK)$, while the explicit-$\boldsymbol{\Psi}$ cost is nearly $K$-independent for $K\ll N$, where its dense $N \times N$ block dominates.
Notably, the mat--vec implementation remains at least as fast over the entire range $25 \leq K \leq 3200$, with the gap narrowing to about $10\%$ near $K \approx N$.
}

\subsection{Empirical Convergence Analysis}
\label{subsec:convergence_exp}

\begin{table}[t]
  \centering
  \caption{\blue{Estimated graph quality per ADMM iteration cap.}}
  \label{tab:convergence}
  \begin{tabular}{rrrrr}
    \toprule
    & \multicolumn{2}{c}{$N{=}50$, $K{=}2000$} &
      \multicolumn{2}{c}{$N{=}100$, $K{=}50$} \\
    \cmidrule(lr){2-3}\cmidrule(lr){4-5}
   iter. cap & Macro-F1 $\uparrow$ & RE $\downarrow$ & Macro-F1 $\uparrow$ & RE $\downarrow$ \\
    \midrule
        30 & 0.783 & 0.217 & 0.100 & 0.394 \\
       100 & 0.865 & 0.130 & \bf{0.138} & \bf{0.378} \\
       300 & \bf{0.873} & \bf{0.088} & 0.123 & 0.427 \\
      1000 & 0.821 & 0.101 & 0.120 & 0.478 \\
      3000 & 0.702 & 0.127 & 0.104 & 0.525 \\
    \midrule
    stopping rule & 0.830 & 0.154 & 0.124 & 0.449 \\
    \bottomrule
  \end{tabular}
\end{table}

\begin{figure}[t]
    \centering
    \subfigure[$N=100$, $K=50$]{\includegraphics[width=0.46\linewidth]{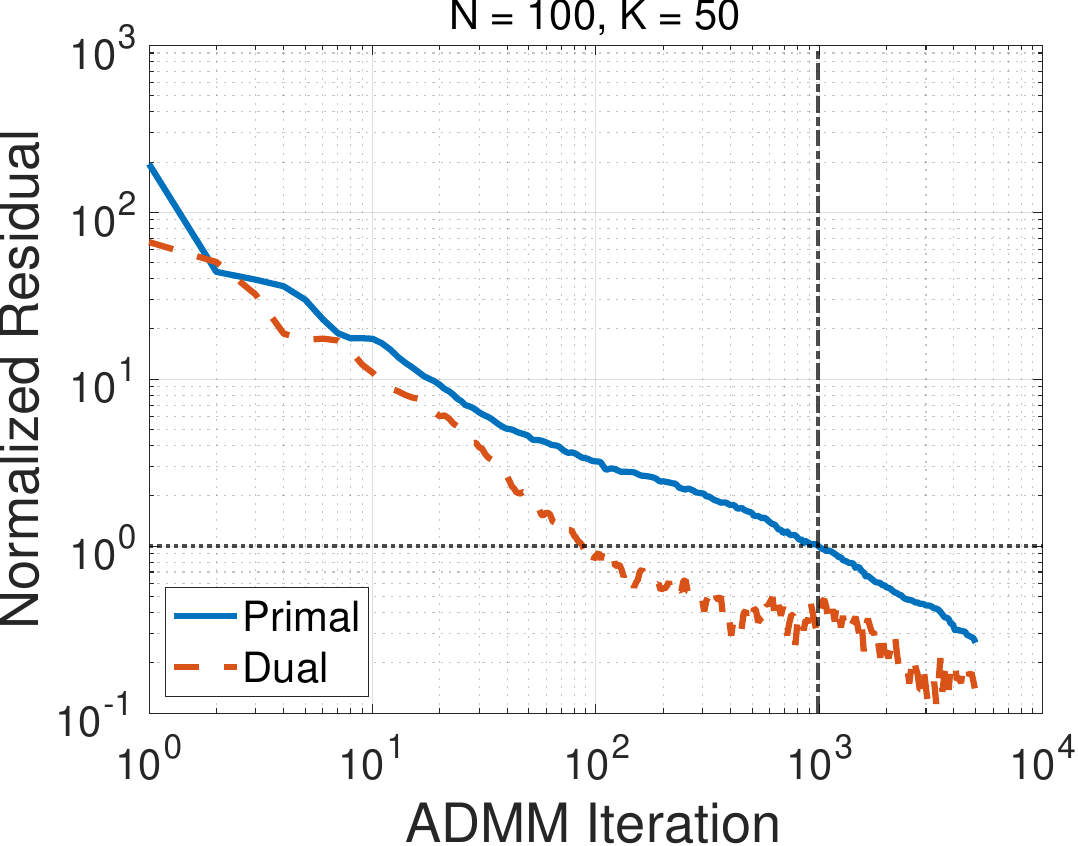}}
    \subfigure[$N=800$, $K=50$]{\includegraphics[width=0.46\linewidth]{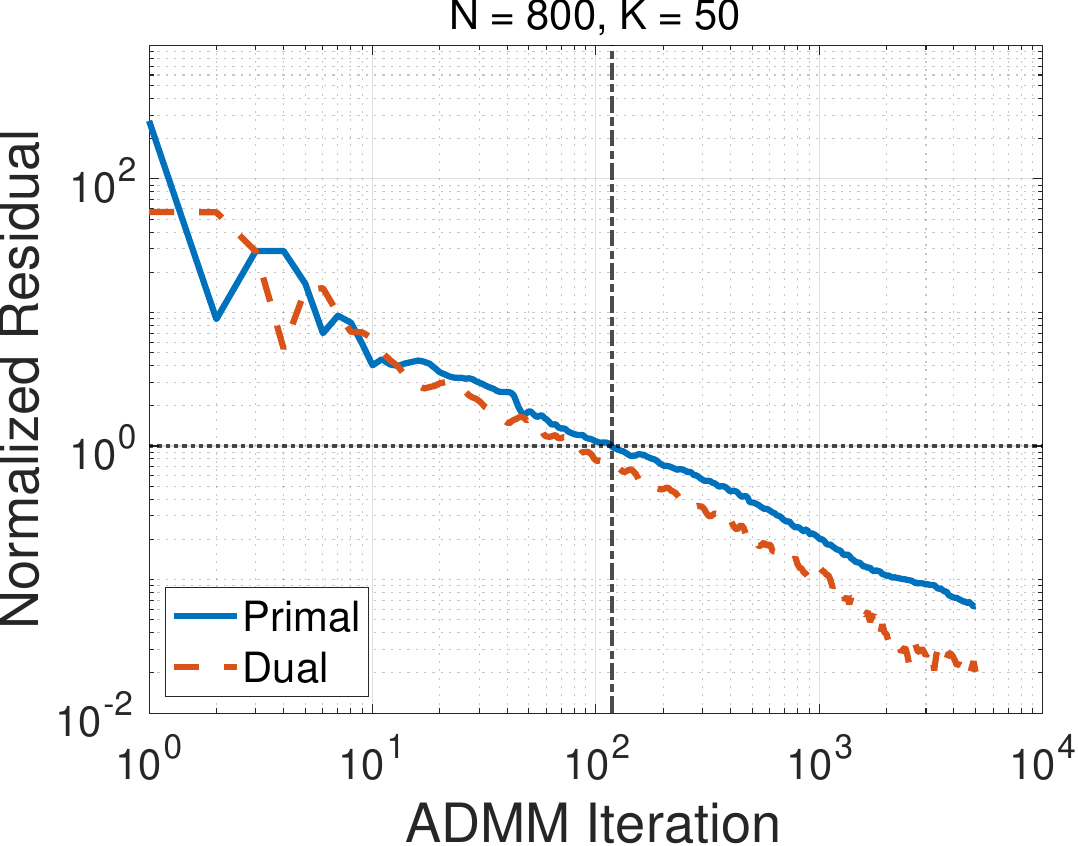}}
  \caption{\blue{Median primal and dual ADMM residuals for column-wise subproblems in the $K\ll N$ regime, normalized by the stopping thresholds (the stopping rule fires when both fall below $1$, dotted line; the vertical line marks the median stopping iteration). The residuals decrease approximately as a power law.}}
  \label{fig:conv_residuals}
\end{figure}

\blue{
We next evaluate the empirical convergence behavior of the proposed algorithm.
The goal is to measure how many ADMM iterations are needed before the learned graph becomes stable.
We run the full proposed algorithm while varying the per-column ADMM iteration cap in $\{30,100,300,1000,3000\}$ with the absolute/relative residual tolerances $10^{-3}$.
The experiment is conducted on the same two synthetic regimes as the noise experiments in Section~\ref{sec:synthetic_noise}: $N=50,K=2000$ and $N=100,K=50$, at $\sigma=0$.
}

\blue{Table~\ref{tab:convergence} summarizes the graph recovery metrics.
The best recovery is reached within $100$--$300$ iterations in both regimes, and further optimization does not improve the graph recovery.
The residual tolerance is reached at median of $53$ iterations per column for $N=50,K=2000$ and $424$ iterations for $N=100,K=50$.
Notably, increasing the iteration cap further does not improve the learned graph quality, since the iterates approach the exact solution of the near-degenerate LP, which is denser than the ground-truth support.
}

\blue{We also plot the ADMM residuals, normalized by their stopping thresholds, in Fig.~\ref{fig:conv_residuals}.
The residuals decrease approximately as a power law.
This behavior is typical for ADMM applied to linear programs, and very small residual tolerances would require many iterations.
For this reason, our implementation combines a moderate residual tolerance with a fixed iteration cap.
The results in Table~\ref{tab:convergence} show that this stopping rule is sufficient for stabilizing the learned graph.
}

\subsection{Graph Signal Restoration on Real-World Datasets}

\begin{figure*}{t}
    \centering
    \subfigure[USS]{\includegraphics[width=0.23\textwidth]{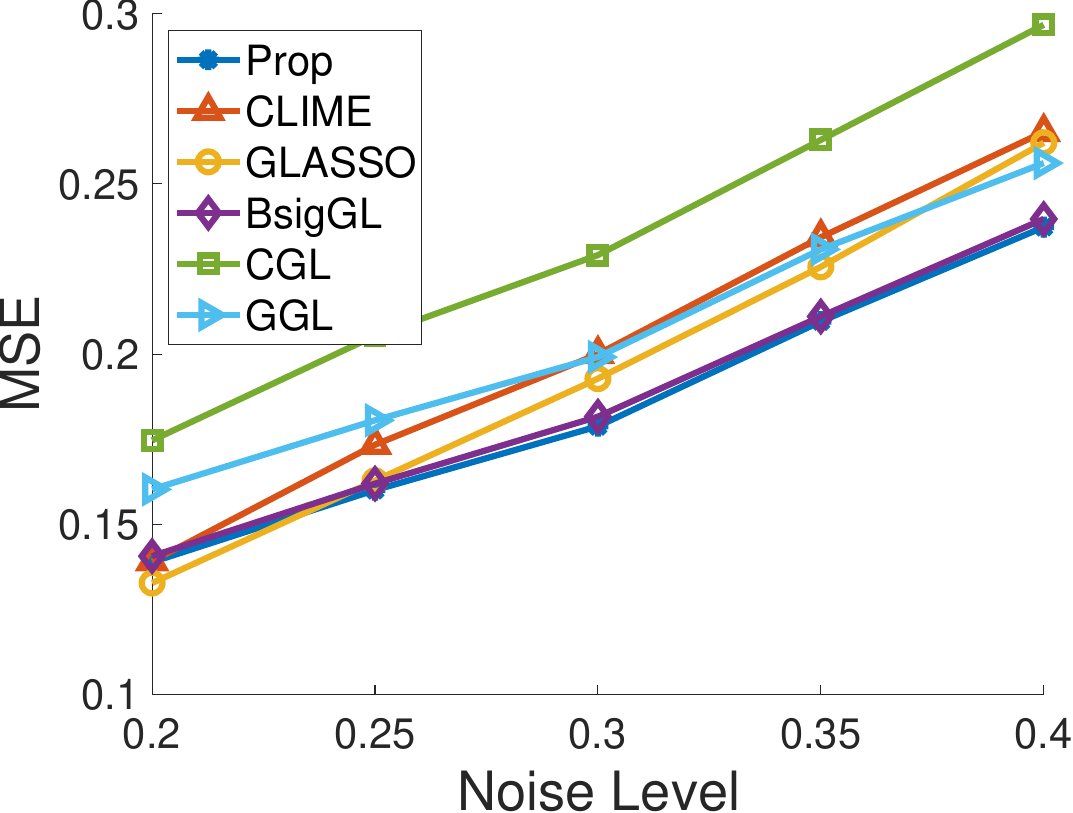}}
    \subfigure[CPV]{\includegraphics[width=0.23\textwidth]{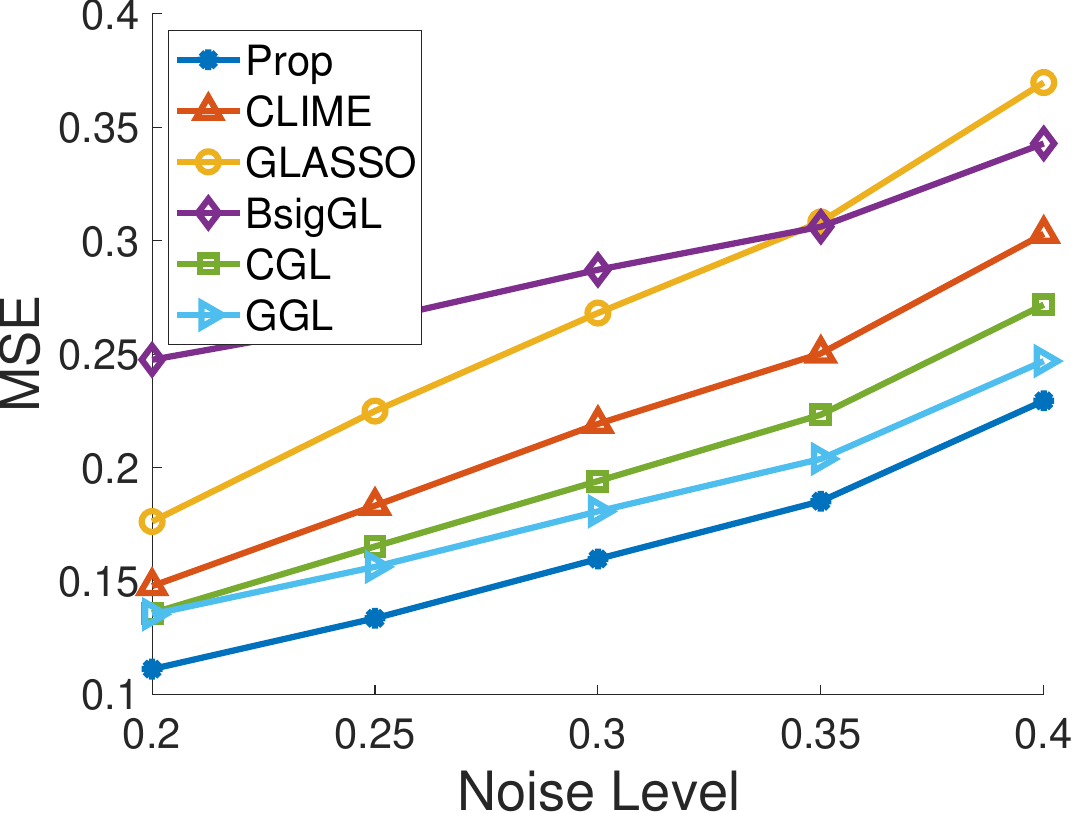}}
    \subfigure[APJ]{\includegraphics[width=0.23\textwidth]{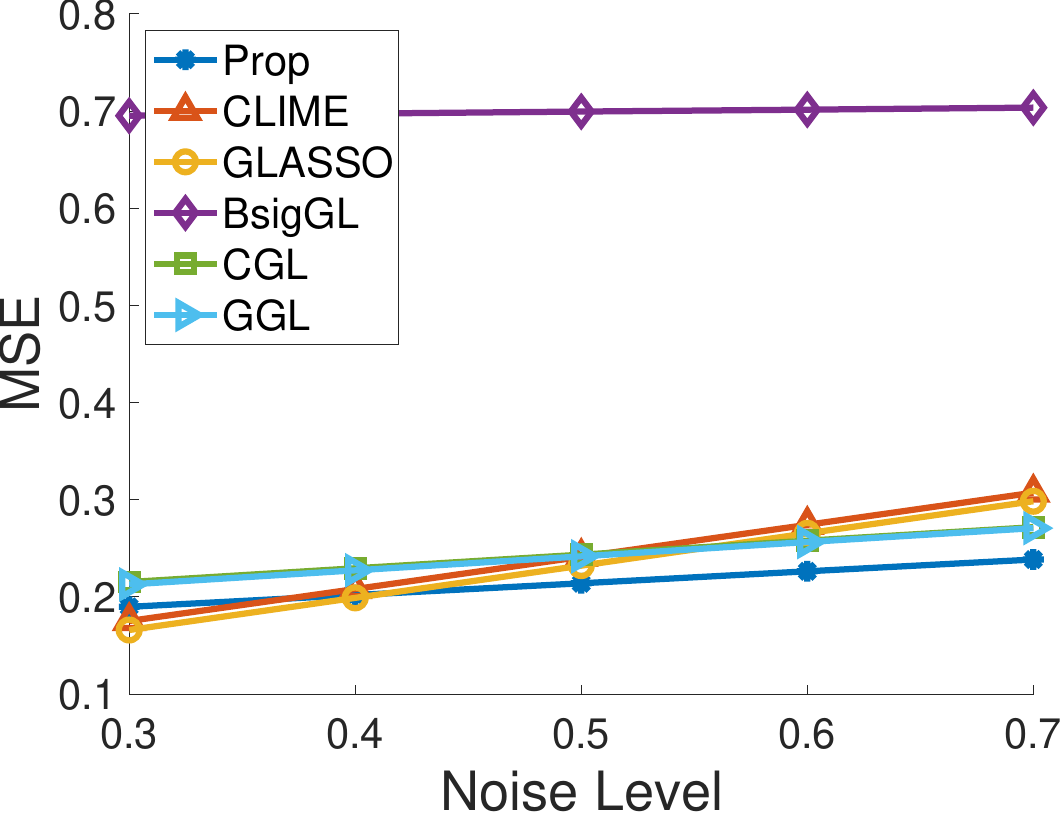}}
    \subfigure[JPS]{\includegraphics[width=0.23\textwidth]{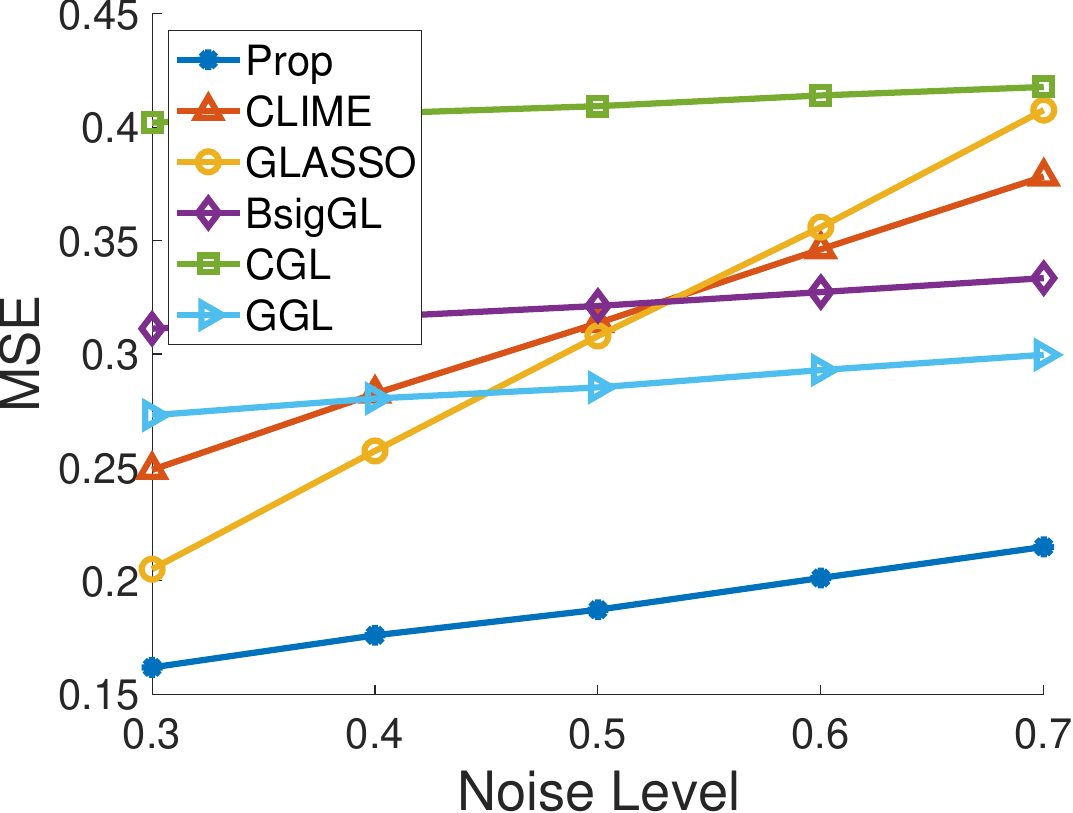}}\vspace{-5pt}\\
    \subfigure[USS]{\includegraphics[width=0.23\textwidth]{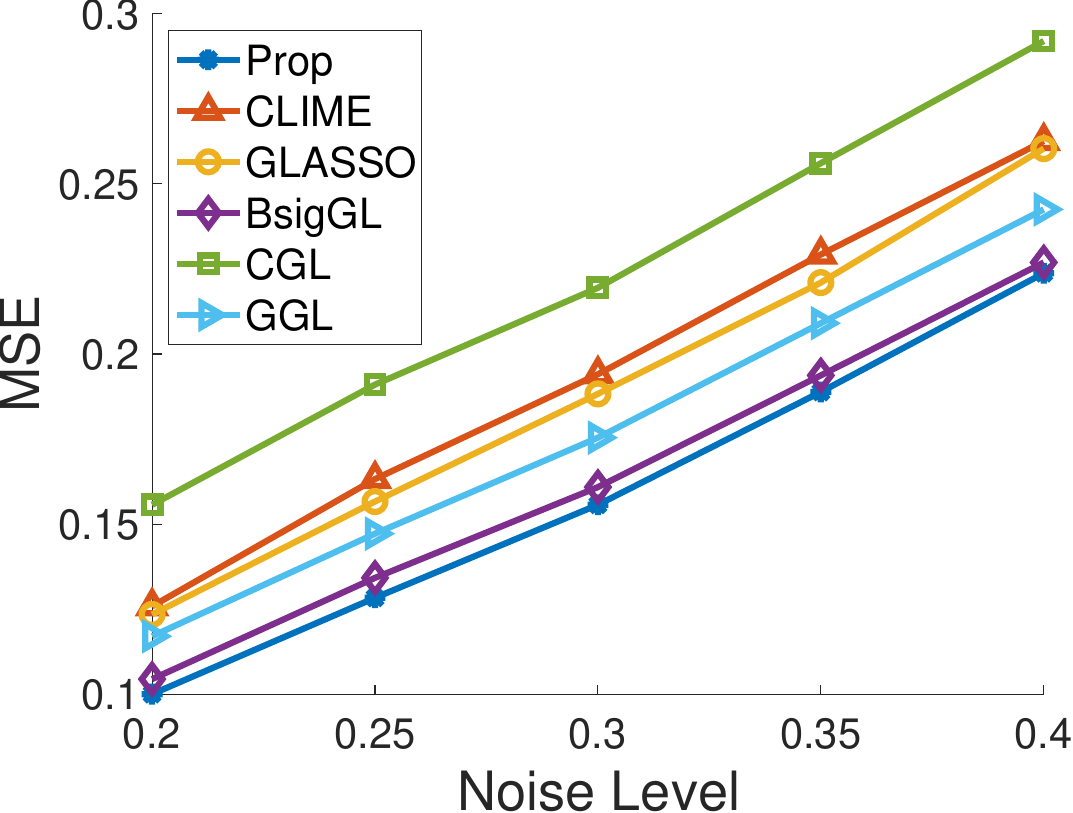}}
    \subfigure[CPV]{\includegraphics[width=0.23\textwidth]{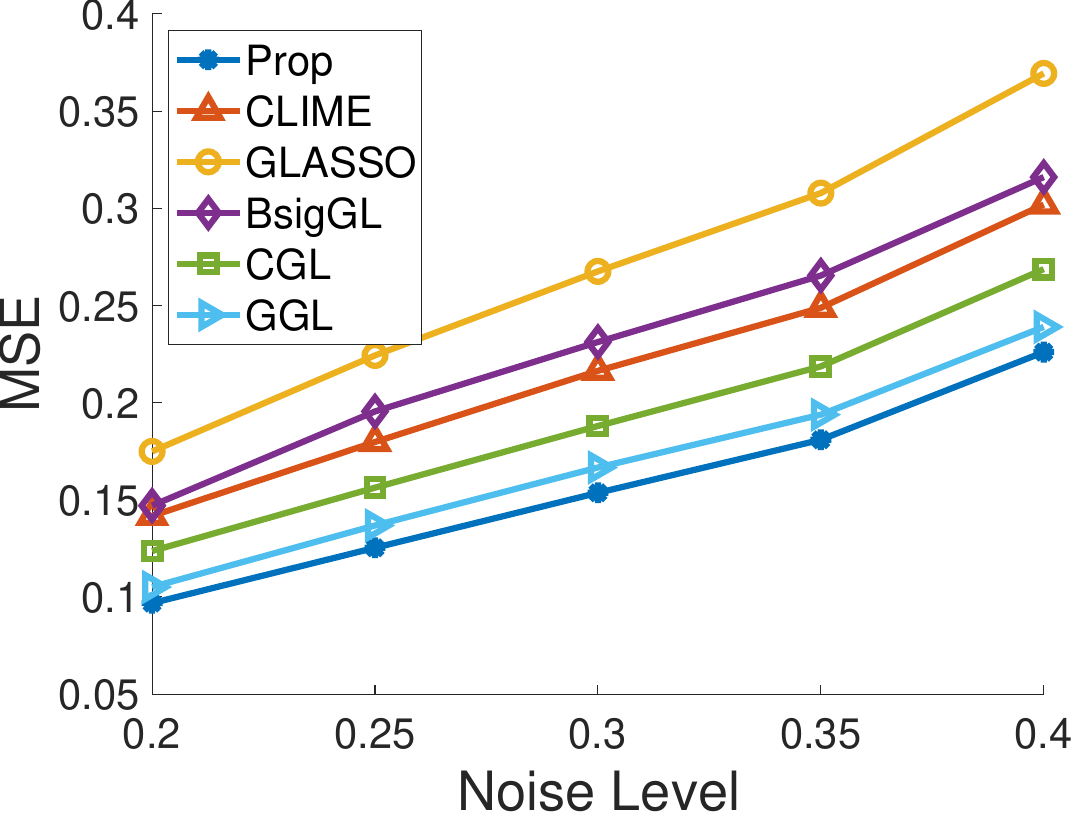}}
    \subfigure[APJ]{\includegraphics[width=0.23\textwidth]{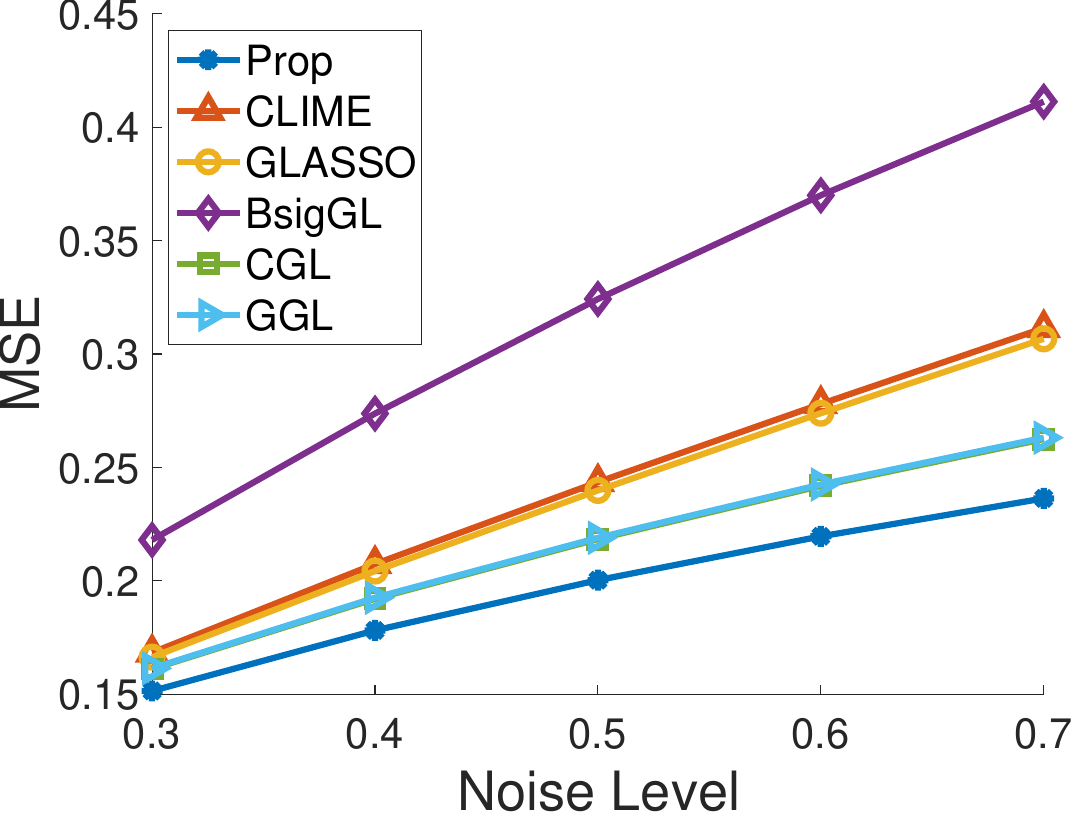}}
    \subfigure[JPS]{\includegraphics[width=0.23\textwidth]{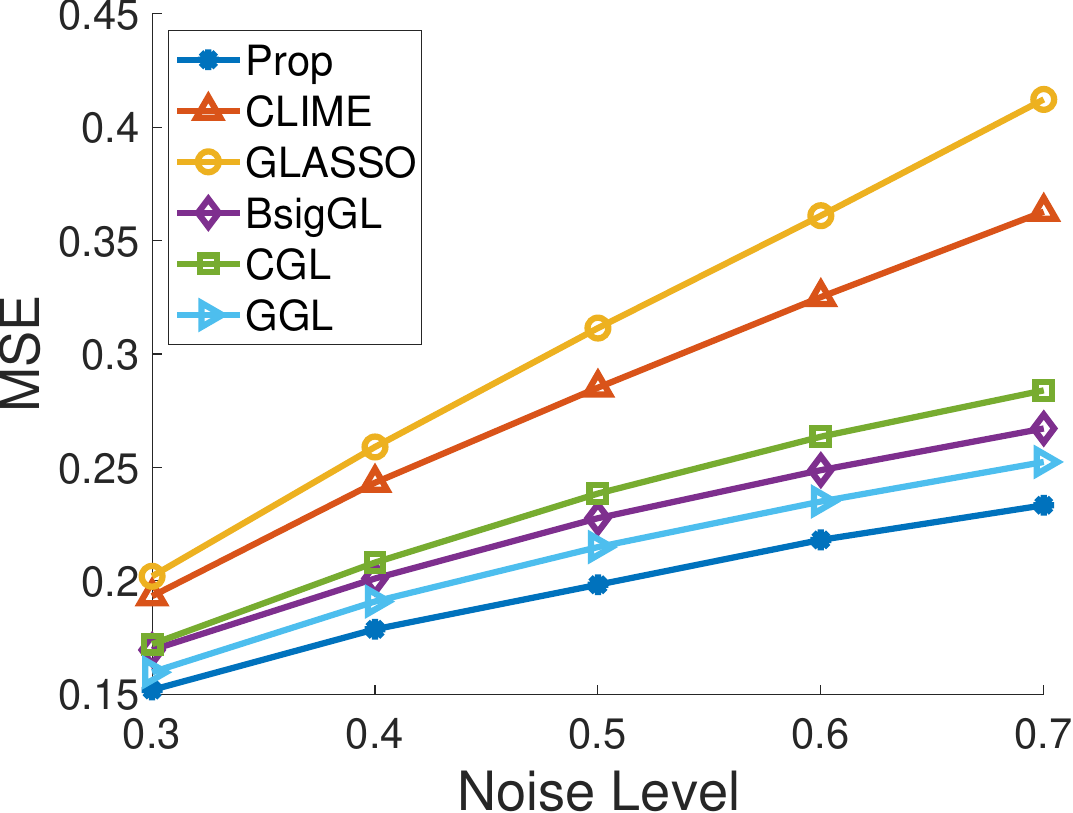}}
    \caption{Results of signal denoising using estimated graph Laplacians. Figures in the top row are the results of graph low-pass filtering (BL), and figures in the bottom row are the results of graph spectral wavelets (SGW). The mean squared errors (MSE) are averaged over 50 independent runs.}\label{fig:denoising}\vspace{-10pt}
\end{figure*}

\begin{table*}[tb]
\caption{MSEs for denoising with graph neural networks.}\vspace{-5pt}
\label{tab:GNN_dn}
\centering
\begin{tabular}{@{}c c@{}}
\begin{tabular}{@{}c c c c c c c c@{}}
\toprule
Dataset & Model & Method & 0.20 & 0.25 & 0.30 & 0.35 & 0.40 \\
\midrule
\multirow{8}{*}{USS}
& \multirow{4}{*}{GCN}
& Prop   & \ul{0.084} & \ul{0.096} & \ul{0.105} & \ul{0.114} & 0.127 \\
& & BsigGL & 0.104 & 0.104 & 0.112 & 0.118 & \ul{0.125} \\
& & CGL    & 0.107 & 0.119 & 0.129 & 0.140 & 0.151 \\
& & GGL    & \bf 0.079 & \bf 0.093 & \bf 0.092 & \bf 0.108 & \bf 0.133 \\
\cmidrule(lr){2-8}
& \multirow{4}{*}{GAT}
& Prop   & \bf 0.104 & \ul{0.128} & \bf 0.131 & \bf 0.139 & \ul{0.159} \\
& & BsigGL & 0.116 & \bf 0.126 & \ul{0.137} & \ul{0.135} & \bf 0.158 \\
& & CGL    & 0.127 & 0.186 & 0.203 & 0.272 & 0.312 \\
& & GGL    & \ul{0.112} & 0.138 & 0.145 & 0.148 & 0.186 \\
\midrule
\multirow{8}{*}{CPV}
& \multirow{4}{*}{GCN}
& Prop   & \bf 0.100 & \bf 0.096 & \bf 0.092 & \bf 0.107 & \bf 0.115 \\
& & BsigGL & 0.123 & 0.121 & 0.157 & 0.180 & 0.203 \\
& & CGL    & 0.166 & 0.177 & 0.202 & 0.222 & 0.247 \\
& & GGL    & \ul{0.109} & \ul{0.108} & \ul{0.117} & \ul{0.128} & \ul{0.136} \\
\cmidrule(lr){2-8}
& \multirow{4}{*}{GAT}
& Prop   & \bf 0.122 & \bf 0.146 & \bf 0.159 & \bf 0.181 & \ul{0.220} \\
& & BsigGL & 0.191 & 0.235 & 0.301 & 0.337 & 0.371 \\
& & CGL    & 0.168 & 0.180 & 0.221 & 0.251 & 0.292 \\
& & GGL    & \ul{0.141} & \ul{0.151} & \ul{0.191} & \ul{0.194} & \bf 0.219 \\
\bottomrule
\end{tabular}
&
\begin{tabular}{@{}c c c c c c c c@{}}
\toprule
Dataset & Model & Method & 0.30 & 0.40 & 0.50 & 0.60 & 0.70 \\
\midrule
\multirow{8}{*}{JPS}
& \multirow{4}{*}{GCN}
& Prop   & \bf 0.069 & \ul{0.101} & \bf 0.132 & \bf 0.145 & \bf 0.193 \\
& & BsigGL & 0.100 & 0.138 & 0.183 & 0.201 & 0.237 \\
& & CGL    & 0.103 & 0.132 & 0.139 & 0.177 & 0.208 \\
& & GGL    & \ul{0.072} & \bf 0.091 & \ul{0.123} & \ul{0.160} & \ul{0.208} \\
\cmidrule(lr){2-8}
& \multirow{4}{*}{GAT}
& Prop   & 0.098 & \ul{0.138} & \bf 0.161 & \bf 0.188 & \bf 0.214 \\
& & BsigGL & 0.109 & 0.165 & 0.240 & 0.258 & \ul{0.297} \\
& & CGL    & \ul{0.095} & 0.156 & 0.208 & 0.242 & 0.308 \\
& & GGL    & \bf 0.091 & \bf 0.128 & \ul{0.167} & \ul{0.233} & 0.302 \\
\midrule
\multirow{8}{*}{APJ}
& \multirow{4}{*}{GCN}
& Prop   & \bf 0.097 & \bf 0.084 & \bf 0.097 & \bf 0.126 & \bf 0.143 \\
& & BsigGL & 0.114 & 0.099 & 0.114 & 0.158 & 0.180 \\
& & CGL    & 0.105 & \ul{0.085} & \ul{0.105} & \ul{0.138} & \ul{0.161} \\
& & GGL    & \ul{0.104} & 0.089 & 0.104 & 0.131 & 0.159 \\
\cmidrule(lr){2-8}
& \multirow{4}{*}{GAT}
& Prop   & 0.080 & 0.122 & 0.151 & 0.185 & 0.232 \\
& & BsigGL & 0.094 & 0.135 & 0.220 & 0.288 & 0.395 \\
& & CGL    & \ul{0.071} & \bf 0.093 & \ul{0.127} & \ul{0.151} & \ul{0.200} \\
& & GGL    & \bf 0.059 & \ul{0.093} & \bf 0.123 & \bf 0.156 & \bf 0.187 \\
\bottomrule
\end{tabular}
\end{tabular}
\vspace{-10pt}
\end{table*}

\begin{table*}[tb]
\caption{MSEs for denoising with signed graph convolutional networks.}\vspace{-5pt}
\label{tab:SGCN_dn}
\centering
\begin{tabular}{@{}c c@{}}
\begin{tabular}{@{}c c c c c c c@{}}
\toprule
Dataset & Method & 0.20 & 0.25 & 0.30 & 0.35 & 0.40 \\
\midrule
\multirow{5}{*}{USS}
& Prop    & \textbf{0.093} & \ul{0.105} & \ul{0.132} & \ul{0.158} & \textbf{0.177} \\
& BsigGL  & \ul{0.094} & \textbf{0.098} & \textbf{0.125} & \textbf{0.139} & \ul{0.183} \\
& CGL     & 0.166 & 0.203 & 0.248 & 0.286 & 0.334 \\
& GGL     & 0.112 & 0.129 & 0.160 & 0.187 & 0.242 \\
& CLIME   & 0.110 & 0.156 & 0.205 & 0.211 & 0.265 \\
\midrule
\multirow{5}{*}{CPV}
& Prop    & \textbf{0.081} & \textbf{0.117} & \textbf{0.124} & \textbf{0.172} & \textbf{0.195} \\
& BsigGL  & 0.120 & 0.148 & 0.176 & 0.194 & 0.219 \\
& CGL     & 0.175 & 0.188 & 0.275 & 0.282 & 0.312 \\
& GGL     & \ul{0.111} & 0.141 & \ul{0.165} & 0.186 & \ul{0.211} \\
& CLIME   & 0.155 & \ul{0.140} & 0.196 & \ul{0.182} & 0.212 \\
\bottomrule
\end{tabular}
&
\begin{tabular}{@{}c c c c c c c@{}}
\toprule
Dataset & Method & 0.30 & 0.40 & 0.50 & 0.60 & 0.70 \\
\midrule
\multirow{5}{*}{JPS}
& Prop    & \textbf{0.105} & 0.147 & \ul{0.180} & \ul{0.207} & \textbf{0.254} \\
& BsigGL  & 0.116 & 0.145 & 0.194 & 0.225 & 0.267 \\
& CGL     & 0.114 & \textbf{0.129} & 0.183 & 0.229 & 0.255 \\
& GGL     & 0.112 & \ul{0.131} & \textbf{0.179} & \textbf{0.188} & 0.267 \\
& CLIME   & 0.124 & 0.145 & 0.195 & 0.236 & 0.282 \\
\midrule
\multirow{5}{*}{APJ}
& Prop    & 0.074 & 0.112 & \ul{0.165} & \textbf{0.218} & \textbf{0.267} \\
& BsigGL  & 0.094 & 0.144 & 0.189 & 0.260 & 0.319 \\
& CGL     & \ul{0.067} & \textbf{0.106} & \textbf{0.164} & 0.223 & 0.279 \\
& GGL     & \ul{0.067} & \ul{0.109} & \ul{0.165} & 0.223 & \ul{0.278} \\
& CLIME   & \textbf{0.065} & 0.110 & \ul{0.165} & \ul{0.221} & 0.300 \\
\bottomrule
\end{tabular}
\end{tabular}
\vspace{-10pt}
\end{table*}

We next evaluate the proposed balanced graph learning method on real-world datasets.
Since we do not have the ground-truth graphs in this setting, we demonstrate the performance via graph signal denoising and interpolation.
For both tasks, we employ graph filters and GNNs, which require a positive graph kernel as input, to show that they can be reused on balanced signed graphs. 

For competing methods, we estimated two signed graphs using GLASSO and CLIME, then constructed a balanced signed graph using \textit{Greed} \cite{Dinesh2025} as in the previous subsection. 
We also performed balanced signature graph learning (BsigGL) \cite{matz23}.
We also compared the performance against positive graphs. 
For positive graph learning, we followed two different formulations in \cite{egilmez_graph_2017}.
The first estimated a combinatorial graph Laplacian (CGL): 
\begin{equation}
\label{eq:CGL}
    \min_{\mathbf{L}} \quad \mathrm{Tr}(\mathbf{L} \mathbf{K}) - \log |\mathbf{L}|\ \quad \text{subject to} \quad \mathbf{L} \in \mathcal{L}_c(\mathbf{A}).
\end{equation}
The second estimated generalized graph Laplacian (GGL):
\begin{equation}
        \min_{\mathbf{L}} \quad \mathrm{Tr}(\mathbf{L} \mathbf{K}) - \log |\mathbf{L}|\ \quad \text{subject to} \quad \mathbf{L} \in \mathcal{L}_g(\mathbf{A})
\end{equation}
where $\mathbf{K}=\mathbf{C}+\mathbf{F}$ is the regularization matrix defined as $\mathbf{F}=\alpha\mathbf{2(\mathbf{I-11}^\mathsf{T})}$, $\mathbf{A}$ is a connectivity matrix, and $\mathcal{L}_c$ and $\mathcal{L}_g$ are feasible sets for CGLs and GGLs, respectively.
Here we set $\mathbf{A}= \mathbf{11}^\mathsf{T}-\mathbf{I}$ because the true connectivity is unknown.

\subsubsection{Datasets} The specifications of the real-world datasets used in this experiment are listed as follows.
\begin{enumerate}
    \item \textit{United States Senate\footnote{https://www.congress.gov/roll-call-votes} (USS, Ternary):} The dataset consists of US Senate voting records from 115th and 116th Congress in 2005-2021 and contains voting records of 100 senators who voted in 1320 elections. 

    \item\textit{Canadian Parliament Voting Records\footnote{https://www.ourcommons.ca/members/en/votes} (CPV, Ternary):} The dataset consists of Canadian Parliament voting records from the 38th parliament in 2005 and contains voting records of 340 constituencies voted in 3154 elections. 

    \item\textit{Hourly Air Pressures in Japan\footnote{https://www.data.jma.go.jp/stats/etrn/index.php} (APJ, Real):} This dataset consists of hourly air pressure records from $48$ weather stations in Japan from March 2022 to May 2022. The number of observations is $K = 2016$. 

    \item\textit{Monthly Japanese Industrial Sales\footnote{https://www.e-stat.go.jp/} (JPS, Real):} This dataset consists of monthly industrial sales from $29$ industries in Japan from December 2020 to December 2023. The total number of observations is $K = 37$. We normalized the sales values of each industry and ignored any industry that had a near-zero correlation coefficient (less than $0.01$) with any other industry.
\end{enumerate}

For the two voting datasets, USS and CPV, the votes were recorded as $-1$ for \textit{no} and $1$ for \textit{yes} and $0$ for \textit{abstain/absent}. We removed any sample with all \textit{yes} or \textit{no} votes, or with more than $50\%$ \textit{abstain/absent} to emphasize the polarity in the dataset. We also ignored constituencies with less than $800$ voting records for the CPV dataset. For the APJ dataset, the data were normalized to be zero-mean and unit variance to emphasize the spatial variance.

For all real-world datasets, we randomly selected $90\%$ of the sample to compute the sample covariance matrix as \blue{$\mathbf{C} = \frac{1}{K-1}\mathbf{XX}^\mathsf{T}+\xi\mathbf{I}$} where $\xi$ is a small constant to ensure $\mathbf{C}$ is non-singular. 
Then, the remaining $10\%$ were used for restoration tasks. 
For the JPS dataset, we only had $K_\text{sample}=4$ test samples, so we generated $10$ different noise realizations to make sure all datasets had at least $K_\text{sample}>N/2$ test samples.

\subsubsection{Signal Denoising Settings}

The test data in the real-valued datasets are contaminated with an additive white Gaussian noise (AWGN) with $\sigma = \{0.3,0.4,0.5,0.6,0.7\}$. 
For the ternary-valued datasets (USS and CPV), we dropped signal values of $n_p N$ nodes from the test data, where $n_p=\{0.2,0.25,0.3,0.35,0.4\}$ to represent a Bernoulli noise.

For positive graph signal denoising, we employed a bandlimited graph low-pass filter (BL), spectral graph wavelets (SGW) \cite{hammond2011wavelets}, graph convolutional net (GCN)\cite{kipf2017semi}, and graph attention net (GAT) \cite{velickovic2018graph}. 
The learned balanced signed graph Laplacian $\mathcal{L}^b$ was similarity-transformed (via matrix $\mathbf{T}$ in \eqref{eq:matrixT}) to a positive graph Laplacian $\mathcal{L}^+$ as a kernel for the denoising algorithms. 
Each observed noisy signal was also similarity-transformed via $\mathbf{T}$ for processing on positive graph $\mathcal{G}^+$.
For BL, we used a filter that preserves the low-frequency band $[0,0.3 \lambda_\text{max}]$, where $\lambda_\text{max}$ was the maximum eigenvalue of the graph Laplacian. 
For SGW, we designed the Mexican hat wavelet filter bank for range $[0, \lambda_\text{max}]$. 
The number of frames was set to $7$. 
For GCN and GAT, we followed the untrained GNN implementation in \cite{rey_2022_untrained} for signal denoising on positive graphs.
\blue{We also evaluated a signed graph convolutional network (SGCN) \cite{derr2018signed}.
For SGCN, signed graph learning methods, including the proposed method, BsigGL, and CLIME, are used directly as signed graph inputs.
For positive graph learning methods, such as CGL and GGL, there are no negative edges.
In this case, the signed convolution reduces to the ordinary positive-edge case.}

\subsubsection{Signal Interpolation Settings}
\begin{figure*}
    \centering
    \subfigure[USS]{\includegraphics[width=0.24\textwidth]{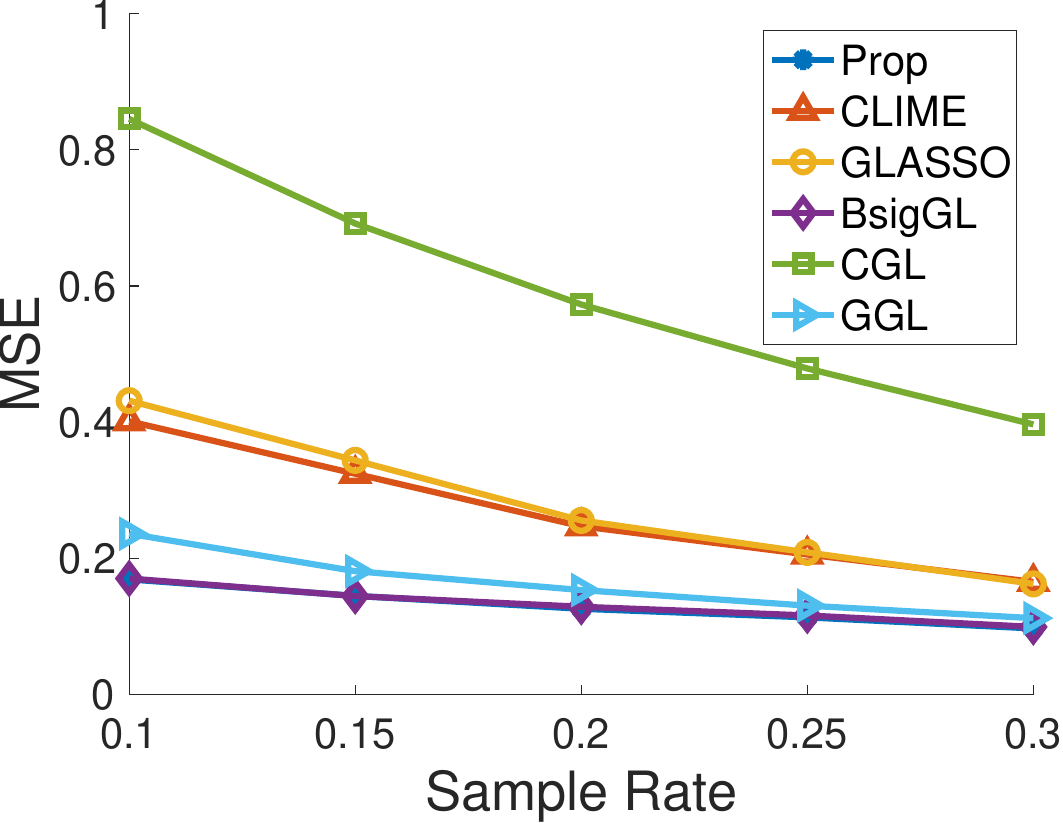}}
    \subfigure[CPV]{\includegraphics[width=0.24\textwidth]{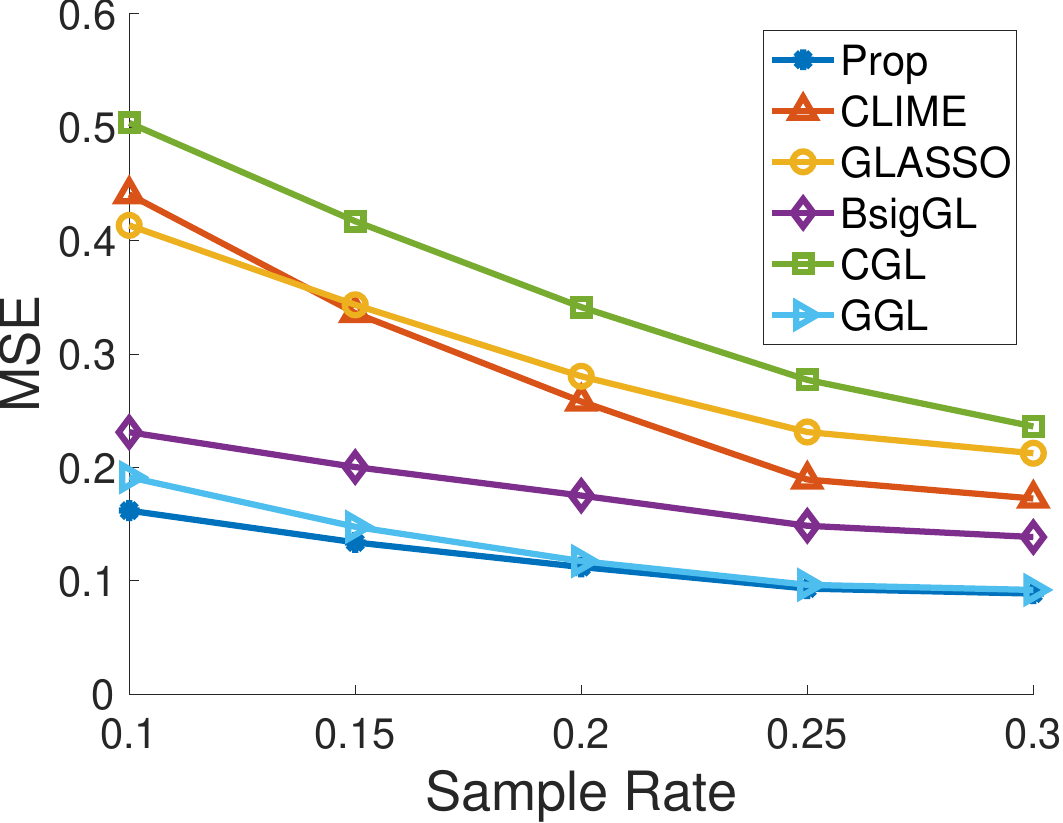}}
    \subfigure[JPS]{\includegraphics[width=0.24\textwidth]{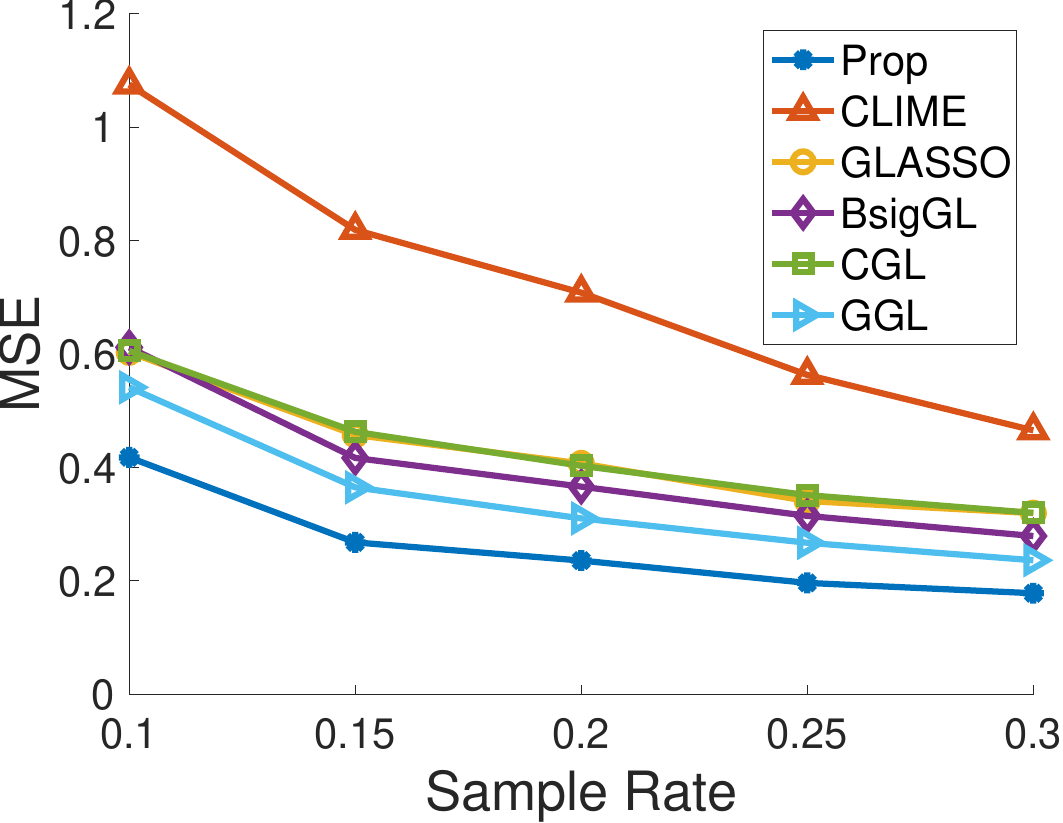}}
    \subfigure[APJ]{\includegraphics[width=0.24\textwidth]{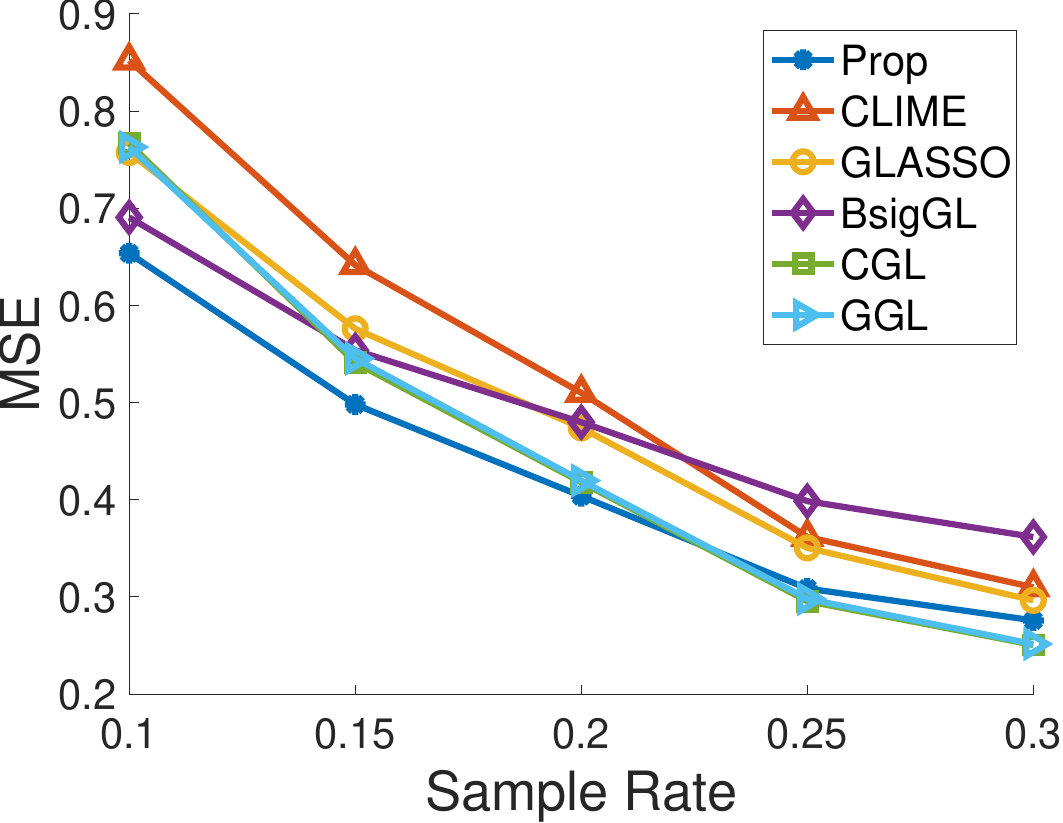}}
    \caption{Results of signal interpolation using estimated graph Laplacians. The mean squared errors (MSE) are averaged over 50 independent runs. }\label{fig:interpolation} \vspace{-10pt}
\end{figure*}

The test data were interpolated from a randomly sampled subset with sample rates $r=\{0.10,0.15,0.20,0.25,0.30\}$.
For signal interpolation, we followed the formulation in \cite{pesenson_variational_2009}. Specifically, \cite{pesenson_variational_2009} requires a positive graph kernel and interpolates missing samples using an upsampling/low-pass filtering procedure. 
Here, the graph low-pass filter with a cutoff frequency of $\lambda_\text{cut}=0.3\lambda_\text{max}$ is defined as $f((1-\lambda_i)/\lambda_\text{cut})$ where
\begin{equation}
    f(x) = \begin{cases}
    0&\text{if}\quad x\le0,\\
    \frac{\exp(-1/x)}{\exp(-1/x)+\exp(-1/(1-x))} &\text{if}\quad x\in(0,1),\\
    1 &\text{if}\quad x\ge 1.
    \end{cases}
\end{equation}
Again, the learned balanced graph Laplacian was similarity-transformed into a positive graph before being used as a kernel for the interpolation algorithm.

\subsubsection{Results}
Figs.~\ref{fig:denoising} and \ref{fig:interpolation} show the mean squared errors (MSE) of the denoised and interpolated signals by filtering-based methods, respectively. The denoising results with GNNs are also presented in Table~\ref{tab:GNN_dn}, where the best results are shown in bold, and the second-best results are underlined.

From Fig.~\ref{fig:denoising}, we observe that the graph learned with the proposed method consistently achieves the lowest MSEs in most cases. 
While BsigGL is comparable to the proposed method for the USS data, its MSEs are larger than the other methods for the other datasets.

\blue{
Table~\ref{tab:GNN_dn} shows that the proposed method consistently performs well with GCN and GAT.
Table~\ref{tab:SGCN_dn} shows the results with SGCN, where the signed graphs are used directly.
The proposed method achieves the best or second-best MSE in most settings.
The advantage is especially clear at higher noise levels.
This suggests that the learned balanced signed graph is useful not only after transformation to a positive graph, but also as an input to signed GNN models.
}

The results on the positive graphs and the balanced signed graphs for the USS and APJ datasets are sometimes similar.
This implies that the nodes in the USS and APJ datasets tend to form separate clusters according to their polarity. This behavior is intuitive: in the USS dataset, senators from opposing parties often cast opposing votes, which naturally induces polarization. Similarly, in the APJ dataset, springtime pressure systems in Japan typically propagate from west to east, leading to spatial separation between high and low pressure regions. In both cases, the dominant structure can be effectively captured by clusters of positively correlated nodes. Consequently, a simple positive graph filter can sufficiently recover the signals, resulting in performance comparable to that of the proposed approach. 
In contrast, the cluster structures in the CPV and JPS datasets are not as clear as those in the USS and APJ datasets. The Canadian parliament consists of five distinct political groups, including parties that occupy middle positions between liberal and conservative ideologies, and industries in Japan do not always rise or fall in a polarized manner.

\begin{figure}{tb}
    \centering
    \subfigure[USS]{\includegraphics[width=0.44\linewidth]{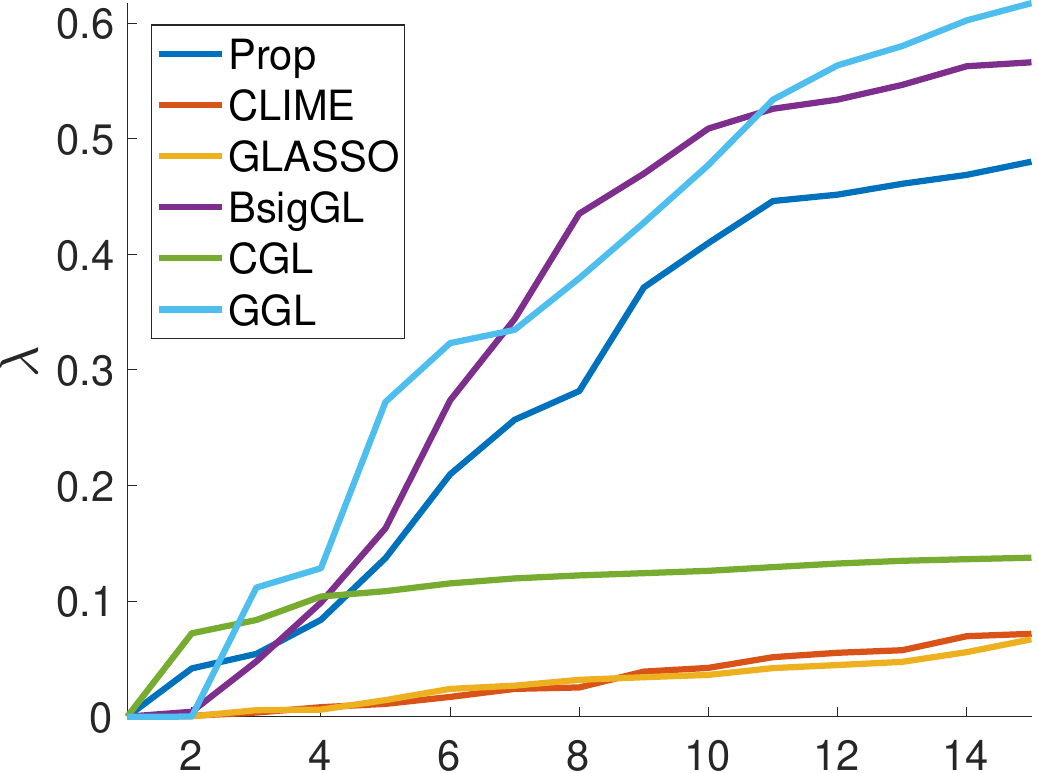}}
    \subfigure[CPV]{\includegraphics[width=0.44\linewidth]{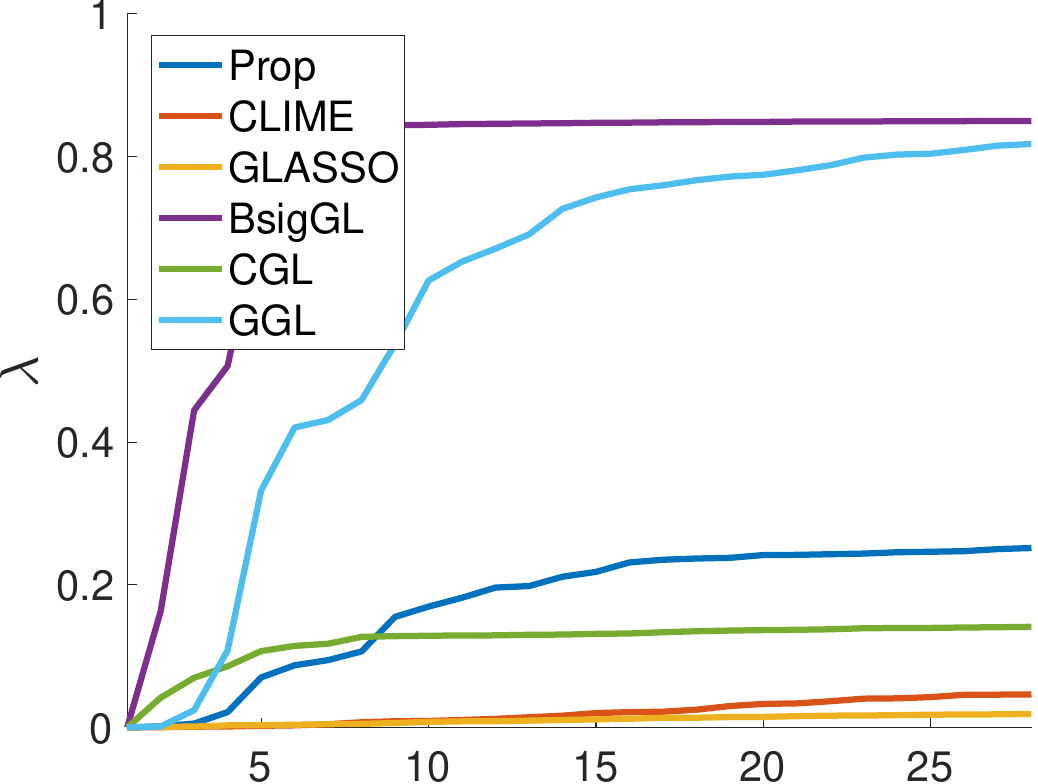}}\\
    \subfigure[JPS]{\includegraphics[width=0.44\linewidth]{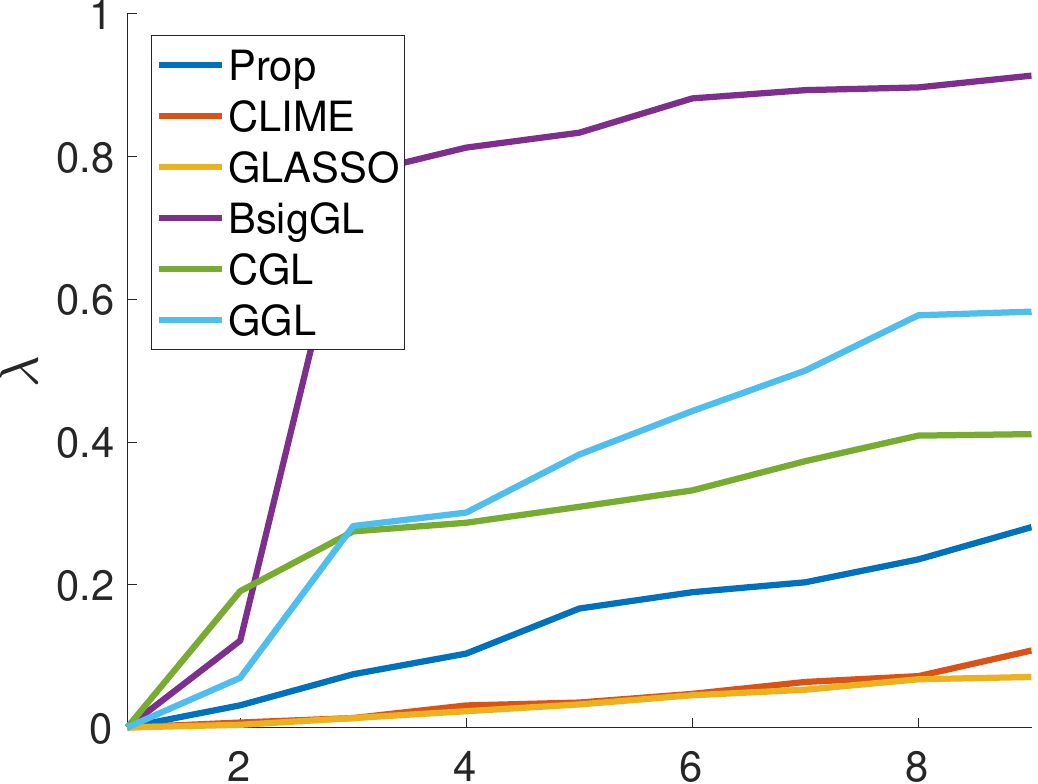}}
    \subfigure[APJ]{\includegraphics[width=0.44\linewidth]{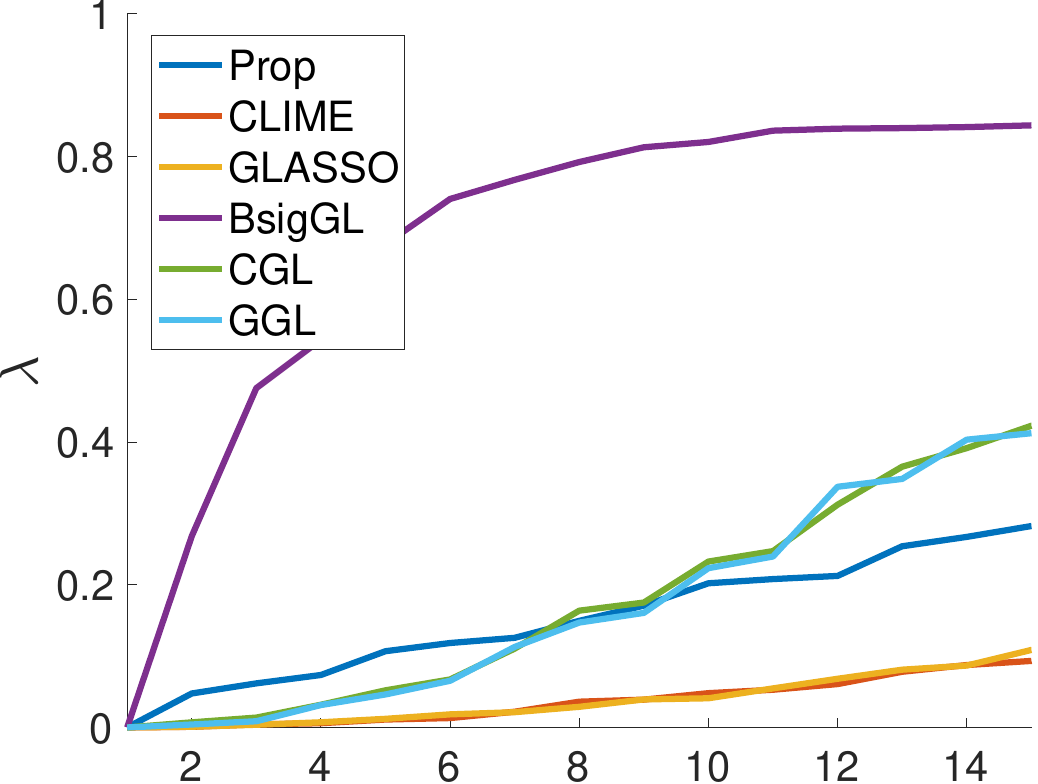}}
    \caption{First $30\%$ eigenvalues of estimated graph Laplacians (rescaled to $[0,1]$). The horizontal axis represents the index of eigenvalues.}
    \label{fig:eigenv}\vspace{-0.2in}
\end{figure}

To investigate structural differences, we compare the spectra of the graphs obtained by the proposed method and those by the alternative methods.
Fig. \ref{fig:eigenv} shows the spectra for the first $30\%$ eigenvalues of learned graphs against their indices.
As can be seen from Figs.~\ref{fig:eigenv}(a) and (d), the spectra of the proposed signed graphs are similar to those of BsigGL and GGL on USS, and are close to the positive graphs on APJ.
This spectral similarity indicates that the graphs are nearly iso-spectral \cite{biggs1993algebraic}, which explains the comparable signal restoration performance for these polarized datasets. Conversely, distinct spectral profiles in CPV and JPS correspond to the mixture of positive and negative relationships, where our method shows advantages.

Overall, our approach demonstrates consistent effectiveness across all datasets, adaptively handling both polarized and complex multi-faceted signal relationships.

\section{Conclusion}
\label{sec:conclude}
We propose an efficient algorithm to learn a balanced signed graph directly from data.
We augment a previous linear programming (LP) based sparse inverse covariance matrix estimation method called CLIME \cite{Cai_aconstrained_2011} with additional linear constraints to enforce graph balance.
For each sparse LP that targets a column $\mathbf{l}_i$ in balanced graph Laplacian $\mathcal{L}^b$, we adapt ADMM to solve $\mathbf{l}_i$ efficiently.
We derive a feasible CLIME parameter $\rho_i$ for each sign-constrained column problem.
\blue{We theoretically prove that the row / column updates are non-increasing in optimization objective, and thus the algorithm terminates in a finite number of iterations.}
Extensive experiments show that our balanced graph learning method enables reuse of spectral filters / GCNs for positive graphs and outperforms previous learning methods.

\blue{The proposed perfectly-balanced graph constraints can suppress meaningful edges when the data-generating graph is non-negligibly frustrated; in this case, the output should be interpreted as a balanced approximation within the proposed model class. Future work will consider soft sign constraints that permit limited inconsistency for near-balanced graph structures.}

\bibliographystyle{IEEEtran}
\bibliography{IEEEabrv,TSP_2024}

\appendix

\subsection{Derivation of Linear System}
\label{append:linSys}

We derive linear system \eqref{eq:linSys} from optimization \eqref{eq:obj_main}.  
\blue{Recall the stacked variable $\mathbf{y} = [\mathbf{m}; \mathbf{r}; \mathbf{q}] \in \mathbb{R}^{4N}$ and the selection matrix $\mathbf{E} \in \{0,1\}^{3N\times 4N}$ defined in Section\;\ref{subsubsec:LPform}.}
Thus,
\begin{align}
\mathbf{B} \left[ \begin{array}{c}
\blue{\mathbf{y}} \\
\blue{\tilde{\mathbf{z}}}
\end{array} \right] & -
\left[ \begin{array}{c}
\mathbf{b} \\
\blue{\mathbf{0}_{3N}}
\end{array} \right] \nonumber\\
&=
\left[ \begin{array}{cc}
\mathbf{A} & \blue{\mathbf{0}_{3N,3N}} \\
\blue{\mathbf{E}} & \blue{- \mathbf{I}_{3N}}
\end{array} \right]
\left[ \begin{array}{c}
\mathbf{y} \\
\blue{\tilde{\mathbf{z}}}
\end{array} \right] -
\left[ \begin{array}{c}
\mathbf{b} \\
\blue{\mathbf{0}_{3N}}
\end{array} \right] \nonumber \\&=
\left[ \begin{array}{c}
\mathbf{A} \mathbf{y} - \mathbf{b} \\
\blue{\mathbf{E} \mathbf{y}} - \blue{\tilde{\mathbf{z}}}
\end{array} \right] .
\end{align}
Thus, the Lagrangian term becomes
\begin{align}
\left[ \boldsymbol{\mu}_1^\top ~~ \boldsymbol{\mu}_2^\top \right]  &
\left[ \begin{array}{c}
\mathbf{A} \mathbf{y} - \mathbf{b} \\
\blue{\mathbf{E} \mathbf{y}} - \blue{\tilde{\mathbf{z}}}
\end{array} \right] \nonumber \\&=
\boldsymbol{\mu}_1^\top \mathbf{A} \mathbf{y} - \boldsymbol{\mu}_1^\top \mathbf{b} + \boldsymbol{\mu}_2^\top \blue{\mathbf{E}} \mathbf{y} - \boldsymbol{\mu}_2^\top \blue{\tilde{\mathbf{z}}}
\nonumber \\
&= \left( \boldsymbol{\mu}_1^\top \mathbf{A} + \boldsymbol{\mu}_2^\top \blue{\mathbf{E}} \right) \mathbf{y} - \boldsymbol{\mu}_1^\top \mathbf{b} - \boldsymbol{\mu}_2^\top \blue{\tilde{\mathbf{z}}}
\end{align}
where $\boldsymbol{\mu}_1 \in \mathbb{R}^{\blue{3N}}$ and $\boldsymbol{\mu}_2 \in \mathbb{R}^{\blue{3N}}$ are multiplier sub-vectors corresponding to \blue{the equality constraints $\mathbf{A}\mathbf{y} = \mathbf{b}$ and the splitting constraints $\mathbf{E}\mathbf{y} = \tilde{\mathbf{z}}$}, respectively.
Similarly, the augmented Lagrangian term is $\gamma/2$ times the following:
\begin{align}
\left( \mathbf{A} \mathbf{y} - \mathbf{b} \right)^\top & \left( \mathbf{A} \mathbf{y} - \mathbf{b} \right)
+ \left( \blue{\mathbf{E} \mathbf{y}} - \blue{\tilde{\mathbf{z}}} \right)^\top
\left( \blue{\mathbf{E} \mathbf{y}} - \blue{\tilde{\mathbf{z}}} \right) \nonumber \\
= \mathbf{y}^\top \mathbf{A}^\top \mathbf{A} & \mathbf{y} - 2 \mathbf{y}^\top \mathbf{A}^\top \mathbf{b} + \mathbf{b}^\top \mathbf{b} \nonumber\\
+ \mathbf{y}^\top &
\blue{\mathbf{E}^\top \mathbf{E}} \, \mathbf{y} - 2 \mathbf{y}^\top \blue{\mathbf{E}^\top} \blue{\tilde{\mathbf{z}}}
+ \blue{\tilde{\mathbf{z}}}^\top \blue{\tilde{\mathbf{z}}} .
\end{align}
The objective \eqref{eq:obj_main} can now be rewritten in terms of $\mathbf{y}$ as

\vspace{-0.05in}
\begin{small}
\begin{align}
\blue{\mathbf{c}^\top} & \mathbf{y} + \frac{\gamma}{2} \mathbf{y}^\top \left( \mathbf{A}^\top \mathbf{A} + \blue{\mathbf{E}^\top \mathbf{E}} \right) \mathbf{y} \nonumber \\
&+\mathbf{y}^\top \left( \mathbf{A}^\top \boldsymbol{\mu}_1 + \blue{\mathbf{E}^\top} \boldsymbol{\mu}_2 + \gamma \mathbf{A}^\top \mathbf{b} - \gamma \blue{\mathbf{E}^\top} \blue{\tilde{\mathbf{z}}}
\right)
\nonumber \\
& - \boldsymbol{\mu}_1^\top \mathbf{b} - \boldsymbol{\mu}_2^\top \blue{\tilde{\mathbf{z}}} + \frac{\gamma}{2} \mathbf{b}^\top \mathbf{b} + \frac{\gamma}{2} \blue{\tilde{\mathbf{z}}}^\top \blue{\tilde{\mathbf{z}}} .
\end{align}
\end{small}
We take the derivative w.r.t. $\mathbf{y}$:
\begin{align}
\blue{\mathbf{c}} &+
\gamma \left( \mathbf{A}^\top \mathbf{A} + \blue{\mathbf{E}^\top \mathbf{E}} \right) \mathbf{y} +
\mathbf{A}^\top \boldsymbol{\mu}_1 \nonumber \\
&+ \blue{\mathbf{E}^\top} \boldsymbol{\mu}_2 + \gamma \mathbf{A}^\top \mathbf{b} - \gamma \blue{\mathbf{E}^\top} \blue{\tilde{\mathbf{z}}} = \mathbf{0}
\nonumber \\
\gamma &\left( \mathbf{A}^\top \mathbf{A} + \blue{\mathbf{E}^\top \mathbf{E}} \right) \mathbf{y} \nonumber \\
&= - \blue{\mathbf{c}}
-
\mathbf{A}^\top \boldsymbol{\mu}_1 - \blue{\mathbf{E}^\top \boldsymbol{\mu}_2} - \gamma \mathbf{A}^\top \mathbf{b} + \gamma \blue{\mathbf{E}^\top \blue{\tilde{\mathbf{z}}}}
\end{align}
which is the same as the linear system \eqref{eq:linSys}.

\subsection{Derivation of Thresholding Operation}
\label{append:threshold}

We derive the solution to optimization \eqref{eq:auxVar}.
Ignoring the first convex but non-smooth term $g(\tilde{\mathbf{q}})$, the remaining two terms in the objective are convex and smooth.
Taking the derivative w.r.t. variable $\tilde{\mathbf{q}}$ and setting it to $\mathbf{0}$,
\begin{align}
- \boldsymbol{\mu}_2 - \gamma \blue{\blue{\mathbf{z}}^{t+1}} + \gamma \tilde{\mathbf{q}}^* &= \blue{\mathbf{0}_{3N}}
\nonumber \\
\tilde{\mathbf{q}}^* &= \blue{\blue{\mathbf{z}}^{t+1}} + \frac{1}{\gamma} \boldsymbol{\mu}_2 .
\label{eq:appendThres}
\end{align}
This solution is valid iff $g(\tilde{\mathbf{q}}^*) = 0$; otherwise the first term $g(\tilde{\mathbf{q}}^*)$ dominates and $\tilde{\mathbf{q}}^* = \blue{\mathbf{0}_{3N}}$.
Given that \eqref{eq:appendThres} can be computed entry-by-entry separately, \eqref{eq:threshold} follows.

\end{document}